\documentclass[sigconf, nonacm]{acmart}

\usepackage{caption}
\usepackage{subcaption}

\usepackage[normalem]{ulem}
\usepackage{color}
\usepackage{booktabs}

\usepackage{multirow}
\usepackage{amsmath}
\usepackage{xspace}
\usepackage{tabularx}

\newcommand{\class}[1]{\ensuremath{\mathsf{#1}\xspace}}

\begin{document}
\title{Benchmarking Badminton Action Recognition with a New Fine-Grained Dataset}

\author {Qi Li}
\affiliation{
    \institution{Auburn University}
    \city{Auburn}
    \country{United States}
    }
\email{qzl0019@auburn.edu}

\author {Tzu-Chen Chiu}
\affiliation{
    \institution{National Central University}
    \city{Taoyuan City}
    \country{Taiwan}
    }
\email{kenchiu0507@gmail.com}

\author {Hsiang-Wei Huang}
\affiliation{
    \institution{National Central University}
    \city{Taoyuan City}
    \country{Taiwan}
    }
\email{wei19750511@gmail.com}

\author {Min-Te Sun}
\affiliation{
    \institution{National Central University}
    \city{Taoyuan City}
    \country{Taiwan}
    }
\email{msun@csie.ncu.edu.tw}

\author {Wei-Shinn Ku}
\affiliation{
    \institution{Auburn University}
    \city{Auburn}
    \country{United States}
    }
\email{weishinn@auburn.edu}



\begin{abstract}

In the dynamic and evolving field of computer vision, action recognition has become a key focus, especially with the advent of sophisticated methodologies like Convolutional Neural Networks (CNNs), Convolutional 3D, Transformer and spatial-temporal feature fusion. These technologies have shown promising results on well-established benchmarks but face unique challenges in real-world applications, particularly in sports analysis, where the precise decomposition of activities and the distinction of subtly different actions are crucial. Existing datasets like UCF101, HMDB51, and Kinetics have offered a diverse range of video data for various scenarios. However, there's an increasing need for fine-grained video datasets that capture detailed categorizations and nuances within broader action categories. In this paper, we introduce the VideoBadminton dataset, which is derived from high-quality badminton footage. Through an exhaustive evaluation of leading methodologies on this dataset, this study aims to advance the field of action recognition, particularly in badminton sports. The introduction of VideoBadminton could not only serve for badminton action recognition but also provide a dataset for recognizing fine-grained actions. The insights gained from these evaluations are expected to catalyze further research in action comprehension, especially within sports contexts.

\end{abstract}

\maketitle  

\begin{figure*}[h!]
    \centering
    \includegraphics[width=\textwidth]{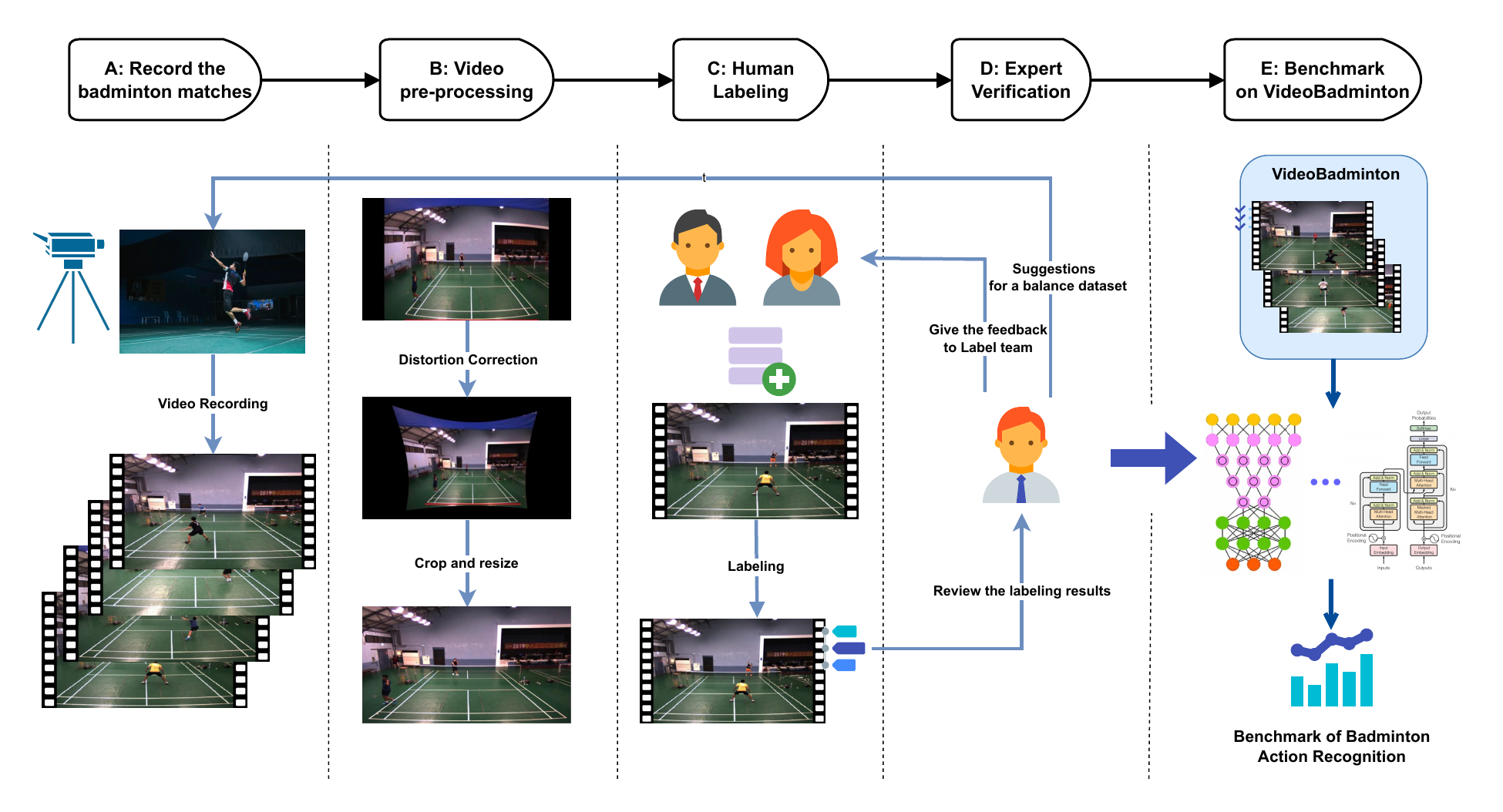}
    \caption{The workflow of creating the VideoBadminton dataset.}
    \label{fig:framework}
    \hfill
\end{figure*}

\section{Introduction}
\label{sec:intro}
Action recognition has become a key focus within the computer vision community, with methodologies leveraging Convolutional Neural Networks (CNNs)~\cite{LeCun1989}, Long Short-Term Memory networks (LSTMs)~\cite{Hochreiter1997}, and the integration of spatial-temporal features~\cite{Karpathy2014} showing notable success on established benchmarks. Nonetheless, real-world applications, particularly in sports analysis, introduce challenges not present in controlled environments~\cite{Poppe2010}. These include decomposing activities into distinct phases and distinguishing between different actions~\cite{soomro2012ucf101}.

The last decade has seen a surge in the development and use of video datasets that have been fundamental to the progression of the computer vision field. Landmark datasets such as UCF101~\cite{soomro2012ucf101}, HMDB51~\cite{kuehne2011hmdb} and Kinetics~\cite{kay2017kinetics} have provided a wealth of video data across a multitude of scenarios and actions. These datasets have been pivotal in advancing the state-of-the-art, inspiring new methodologies~\cite{Simonyan2014}, and enhancing existing ones~\cite{Szegedy2015}. The diverse and extensive content of these datasets has been crucial in training robust models capable of action recognition in various settings and conditions.

Moreover, there has been a shift towards fine-grained video datasets, which aim to capture detailed categorizations within broader action classes, significant in practical applications where the distinction between minor action variations is substantial~\cite{Gkioxari2015}.

Alongside the evolution of video datasets, deep learning methodologies for action recognition have rapidly advanced. CNNs and RNNs, in particular, have taken the lead, utilizing extensive data to extract complex spatial and temporal features from videos~\cite{Donahue2015}. Innovative approaches such as two-stream CNNs~\cite{Simonyan2014} and 3D CNNs~\cite{Ji2013}, along with the combination of CNNs with LSTMs~\cite{Donahue2015} or Transformers~\cite{Vaswani2017}, have set new standards in interpreting video sequences.

To contribute to this field, we introduce VideoBadminton, an innovative dataset curated from high-quality badminton footage. This dataset provides diversity and a finer level of action categorization and requires algorithms that can decode temporal patterns and subtle action differences within continuous activity. This paper presents a comprehensive evaluation of prominent methodologies using the VideoBadminton dataset, offering insights that are expected to stimulate further research in action recognition, particularly in the sports domain.

Our primary contributions are:
\begin{enumerate}
    \item We introduce \textit{VideoBadminton}, a dataset specifically designed for badminton action recognition which is a comprehensive collection of player movements and stroke techniques.
    \item We perform a detailed and systematic evaluation of contemporary state-of-the-art video recognition architectures. This evaluation is not merely an assessment of their performance metrics but also an exploration of their applicability and efficacy in the nuanced context of badminton action recognition.
\end{enumerate}

\section{Related works}

\subsection{Video dataset for action recognition}
In the domain of action recognition, there has been a substantial enhancement in the richness, quality, and variety of datasets for video classification tasks. Seminal datasets such as HMDB51~\cite{kuehne2011hmdb} and UCF101~\cite{soomro2012ucf101} were among the first to augment the variety of action categories and the volume of video content, laying the groundwork for training deep learning models. More recently, the advent of expansive datasets like THUMOS14~\cite{THUMOS14}, Sports-1M~\cite{karpathy2014large}, Kinetics~\cite{kay2017kinetics}, and Moments in Time~\cite{monfort2019moments} have significantly contributed to the advancement of action recognition methodologies. Despite these developments, the primary focus of these datasets remains on coarse-grained recognition. The classification boundaries defined by these datasets are efficient in identifying clear-cut positive instances but deficient in distinguishing between closely related negative actions, which are prevalent in practical settings. Additionally, the entanglement of background context with the actions during the learning process remains a persistent challenge. 

\subsection{Fine-grained video dataset for action recognition}
Efforts have been undertaken to construct datasets geared toward fine-grained action recognition. Notably, the Breakfast and MPII-Cooking 2~\cite{rohrbach2012database} datasets offer annotations for individual stages of various cooking processes. In the context of the Breakfast dataset, coarse actions (e.g., Juice) are dissected into granular action units (e.g., cut orange), while in MPII-Cooking 2, the verbal segments are characterized as fine-grained categories (e.g., cut as an element of the action cutting onion). The Something-Something~\cite{goyal2017something} dataset incorporates 147 categories of everyday human-object interactions, encompassing actions such as lowering an object and retrieving something from a location. The Diving48~\cite{kanojia2019attentive} dataset is established on 48 intricate diving actions, wherein the labels are amalgamations of four attributes, such as back+15som+15twis+free.

While there have been prior efforts~\cite{davar2011domain},~\cite{teng2011detection},~\cite{careelmont2013badminton},~\cite{ting2014automatic},~\cite{ramasinghe2014recognition},~\cite{shan2015investigation},\\~\cite{ting2016potential},~\cite{chu2017badminton},~\cite{weeratunga2017application} in badminton dataset creation, most have predominantly focused on image-level data rather than video. Notably, two research projects have contributed to establishing badminton video datasets, each with unique features and applications.

The Badminton Olympic dataset, as detailed in~\cite{ghosh2018analysis}, comprises 10 videos sourced from YouTube, each showcasing individual badminton matches with an average duration of one hour. This dataset is enriched with diverse annotations, including: 1. Bounding boxes marking player positions in 1,500 frames. 2. Notations of 751 instances where players score points. 3. Comprehensive annotations delineating time boundaries and stroke labels across 12 stroke categories, such as serves and lobs. ShuttleNet~\cite{wang2022shuttlenet} represents a more recent contribution, encompassing 43,191 meticulously selected video clips. Each clip in this dataset focuses on a particular stroke from 10 distinct categories, including smashes, pushes, and lobs. ShuttleNet extends its utility beyond mere action recognition; it is tailored for stroke prediction, where the model is trained to anticipate the next stroke based on the sequence of preceding strokes during a rally.

Our work, VideoBadminton, however, surpasses these datasets in scope and depth. It incorporates 18 categories of badminton actions, with a total of 7,822 clips spanning 145 minutes of self-recorded footage. It is particularly valuable for research in fine-grained action recognition, offering a rich data source for training and evaluating state-of-the-art video recognition models focusing on the nuanced movements and strokes found in badminton. Additionally, VideoBadminton includes detailed annotations of player locations and the trajectory of the shuttle. Given these features, we posit that VideoBadminton currently stands as the most comprehensive dataset for badminton video analysis, setting a new benchmark in the field. See the comparison in the Table \ref{fig:comparision_datasets}.

\begin{table*}
\caption{Comparison of different video datasets, including annotation category, video samples and class number, total running time, data source, and publication year.}
\begin{tabular}{ccccccc}
\hline
Annotation Category       & Dataset                   & No. of Clips & No. of Classes & Total running time (min) & Datasource    & Year \\ \hline
\multirow{2}{*}{Category} & Kinetics400~\cite{kay2017kinetics}      & 306k          & 400          & 50k                      & Internet      & 2017 \\
                          & SthSthv1~\cite{goyal2017something}        & 108k          & 174          & 7.3k                     & Crowdsourcing & 2017 \\ \hline
           & MPII-Cooking2~\cite{rohrbach2012database}    & 273           & 88           & 1.6k                     & Self-recorded & 2012 \\
                          & Breakfast~\cite{kuehne2014language}        & 433           & 50           & 180                      & Self-recorded & 2014 \\
            Category              & THUMOS14~\cite{THUMOS14}         & 413           & 20           & 1.7k                     & Internet      & 2014 \\
          Temporal                & ActivityNet-1.3~\cite{heilbron2015activitynet}   & 19,994        & 200          & 40k                      & Internet      & 2015 \\
                          & Charades~\cite{sigurdsson2016hollywood}         & 9,848         & 157          & 4.9k                     & Crowdsourcing & 2016 \\
                          & FineGym-1.1~\cite{shao2020finegym}     & 12,818        & 530          & 9.8k                     & Internet      & 2020 \\ \hline
   & UCF-sports~\cite{rodriguez2008action}      & 150           & 10           & 5.3                      & Internet      & 2008 \\
  Category & Epic-kitchens-100~\cite{damen2020rescaling} & 700           & 4,053        & 6k                       & Self-recorded & 2020 \\
        Temporal                  & ADL~\cite{pirsiavash2012detecting}             & 20            & 18           & 600                      & Crowdsourcing & 2012 \\
Spatial      & J-HMDB~\cite{kuehne2011hmdb}           & 928           & 21           & 22                       & Internet      & 2013 \\
                          & UCF101-24~\cite{soomro2012ucf101}        & 3,207          & 24           & 385                      & Internet      & 2017 \\
                          & VideoBadminton (\textbf{Ours}) & 7,822           & 18        & 145                       & Self-recorded & 2023 \\

\hline
\end{tabular}

\label{fig:comparision_datasets}

\end{table*}

\subsection{Action recognition models}

The field of action recognition has witnessed substantial advancements in recent years, driven by the development of diverse models that leverage the increasing computational power and sophistication of machine learning techniques. Pioneering efforts in this domain have primarily focused on leveraging convolutional neural networks (CNNs), which have proven effective in extracting spatial features from individual frames. Seminal models like the two-stream CNN~\cite{Simonyan2014}, as proposed by Simonyan and Zisserman, combine spatial and temporal network streams to capture both appearance and motion information, setting a foundation for subsequent innovations.

Building upon this, the integration of Recurrent Neural Networks (RNNs) and Long Short-Term Memory (LSTM) networks has been explored to better understand temporal dynamics in video sequences better. These models, such as the work by~\cite{Donahue2015}, aim to capture the temporal evolution of actions, offering a more nuanced understanding compared to frame-based approaches.

More recent developments have seen the emergence of 3D Convolutional Neural Networks (3D CNNs), which directly process spatio-temporal information in video data.~\cite{carreira2017quo}'s introduction of the I3D model, which extends the concept of 2D CNNs to 3D, has been particularly influential in capturing more complex motion patterns.

Furthermore, attention mechanisms and transformer-based models, like the Vision Transformer (ViT) for action recognition, have gained traction. By focusing on salient features in both spatial and temporal dimensions, these models offer promising avenues for enhancing the accuracy and efficiency of action recognition systems.

The landscape of action recognition models is thus characterized by a continuous evolution from basic frame-level feature extraction to complex spatio-temporal and attention-based models, each contributing to the progress of this field towards more accurate and efficient recognition of human actions in videos.

\section{The VideoBadminton dataset}

In this section, we delve into the comprehensive process of constructing the VideoBadminton dataset. This detailed exposition encompasses the phases of data collection, meticulous data cleaning, preprocessing, and the annotation process. To provide a more granular understanding, we will illustrate the annotation tools and offer examples of video samples. In our commitment to fostering research and collaboration in this field, we will make the complete dataset and the associated codes available for research purposes. This initiative aims to facilitate further advancements and explorations in badminton action recognition and related areas.

\subsection{Row data collection}
In our pursuit of accuracy and comprehensiveness within the VideoBadminton dataset, we meticulously curated practice game videos featuring 19 adept players from National Central University's badminton school team. This cohort included 15 male and 4 female players, each exhibiting a high level of skill closely paralleling professional standards. Their advanced proficiency and tactical prowess in badminton provided a rich reference point for our analytical endeavors. The selection of these players was strategic, enabling us to encapsulate a wide spectrum of badminton actions.

\begin{figure*}[h!]
    \centering
    \begin{subfigure}[b]{0.9\textwidth}
        \includegraphics[width=\textwidth]{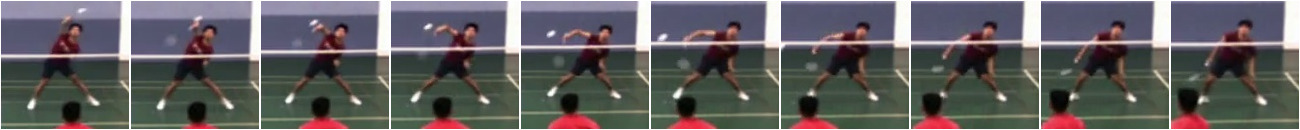}
        \caption{Class of \class{Tap Smash}.}
 
    \end{subfigure}
    \hfill
    \begin{subfigure}[b]{0.9\textwidth}
        \includegraphics[width=\textwidth]{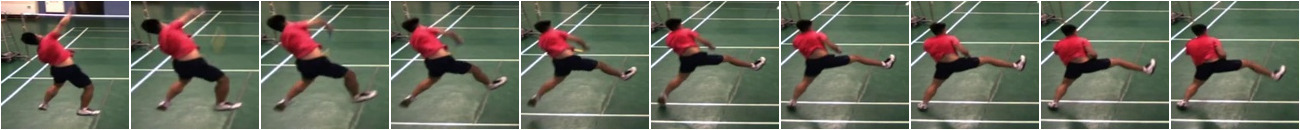}
        \caption{Class of \class{Smash}.}

    \end{subfigure}
    \begin{subfigure}[b]{0.9\textwidth}
        \includegraphics[width=\textwidth]{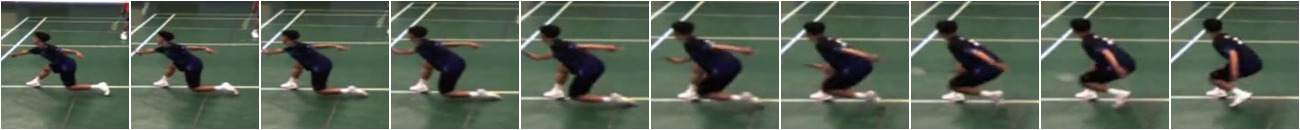}
        \caption{Class of {Defensive clear}.}
 
    \end{subfigure}
    \hfill
    \begin{subfigure}[b]{0.9\textwidth}
        \includegraphics[width=\textwidth]{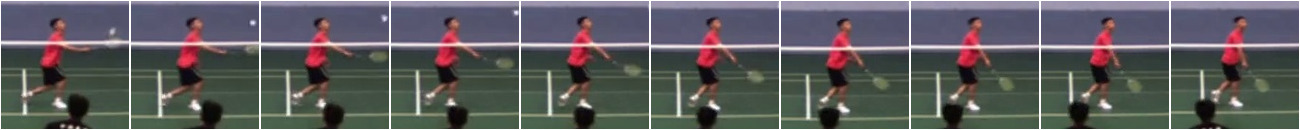}
        \caption{Class of \class{Drop shot}.}
 
    \end{subfigure}
        \begin{subfigure}[b]{0.9\textwidth}
        \includegraphics[width=\textwidth]{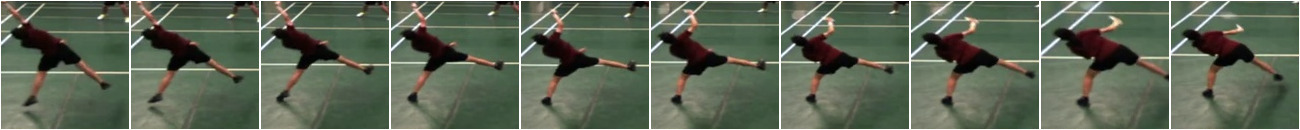}
        \caption{Class of \class{Clear}.}

    \end{subfigure}
    \hfill
    \begin{subfigure}[b]{0.9\textwidth}
        \includegraphics[width=\textwidth]{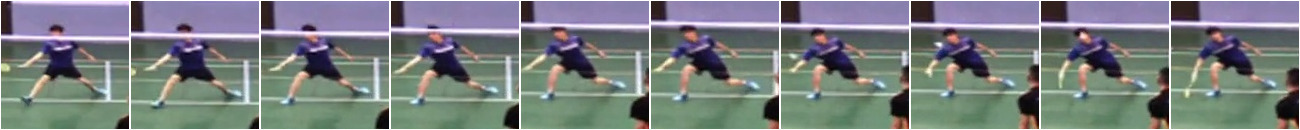}
        \caption{Class of \class{Cross-Court flight}.}
    \end{subfigure}
    \caption{The key frame samples from various badminton classes in VideoBadminton.}
    \label{fig:action_samples}
\end{figure*}

The selected action classes for the badminton dataset aligned with the Badminton World Federation (BWF) standards. These classes are integral for training video recognition models in badminton and include 18 distinct stroke types such as \class{Short~Serve}, \class{Cross-Court~Flight}, \class{Lift}, \class{Tap~Smash}, \class{Block}, \class{Drop~Shot}, \class{Push~Shot}, \class{Transitional~Slice}, \class{Cut}, \class{Rush~Shot}, \class{Defensive~Clear}, \class{Defensive~Drive}, \class{Clear}, \class{Long~Serve}, \class{Smash}, \class{Flat~Shot}, \class{Rear~Court~Flat~Drive}, and \class{Short~Flat~Shot}. This comprehensive set of actions (See the samples in Figure \ref{fig:action_samples}) covers badminton's diverse and dynamic nature, providing a robust foundation for modeling and recognizing various movements in the sport.

The collaboration with the school team proved instrumental in the data collection process. It allowed for the systematic recording of practice matches, showcasing a breadth of badminton actions, and also allowed us to document rarer movements and other nuanced aspects of the sport. This symbiotic relationship was pivotal in augmenting the dataset's diversity, ensuring its thoroughness in capturing an array of actions reflective of high-level badminton gameplay. Thus, the resultant dataset is a comprehensive and detailed compilation, critical for studying and understanding badminton actions.

\subsection{Camera settings}
We used a high-quality camera from The Imaging Source during data collection, specifically the model DFK 37AUX273. This camera was equipped with specific parameters to capture detailed video footage of badminton actions. The camera’s resolution was set at 1280*960 pixels to represent the badminton movements clearly. Operating at a frame rate of 60fps, the DFK 37AUX273 captured the fast-paced movements inherent in badminton, resulting in smooth and high-definition video recordings. The XRGB video format was selected to preserve essential color information during data collection.

\begin{figure}[ht]
\centering
\includegraphics[width=0.45\textwidth]{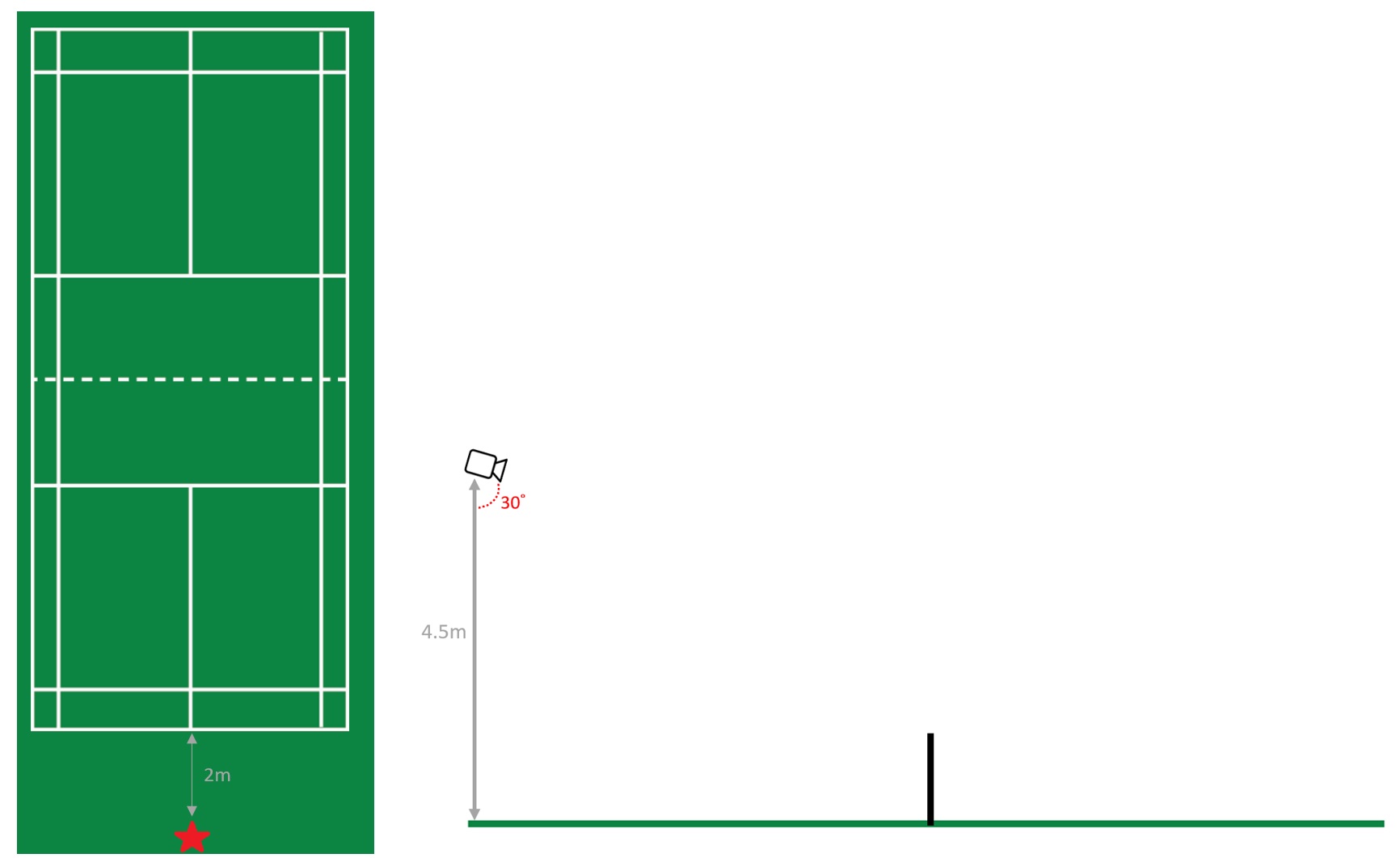}
\caption{The camera setting for recording badminton actions. The camera was placed 2 meters behind the court's baseline and elevated to 4.5 meters, tilted at 30 degrees to capture the actions with minimal distortion.}
\label{fig:camera_fig}
\end{figure}

Strategic camera placement was considered essential. The camera was positioned approximately 2 meters behind the baseline of the badminton court and elevated to a height of approximately 4.5 meters (see Figure \ref{fig:camera_fig}). Tilted at a 30-degree angle, the DFK 37AUX273's lens offered an optimal field of view, reducing distortions and facilitating precise documentation of the various badminton actions. Utilizing this specific model from The Imaging Source allowed for the accurate capture of the intricate movements and techniques of the players, contributing to the comprehensive and detailed dataset.

\subsection{Data preprocessing}
\begin{figure}[ht]
\centering
\includegraphics[width=0.45\textwidth]{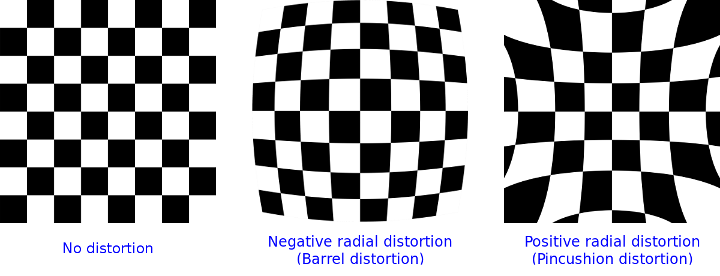}
\caption{The common types of radial distortion. }
\label{fig:distortion_samples}
\end{figure}

We used a wide-angle lens and positioned the camera high up to simulate the broadcast camera used in formal games. However, distortion caused by the wide-angle lens can affect the model's training, resulting in inaccurate court coordinates and deviations in player and shuttlecock positions, leading to errors. This can reduce the availability and accuracy of the data, thereby interfering with the model training results. Therefore, it is necessary to correct the original video. We will use OpenCV to calculate the calibration parameters of the camera used to remove the distortion caused by the lens. The calculation method is to use OpenCV to obtain the 2D coordinates of the chessboard in the 3D world and integrate the information captured from multiple angles of the chessboard image into the calculation, as shown in Figure \ref{fig:distortion_samples}. 

\begin{figure}[t]
	\centering
	\begin{subfigure}[t]{0.8\linewidth}
		\centering
		\includegraphics[width=1.0\linewidth]{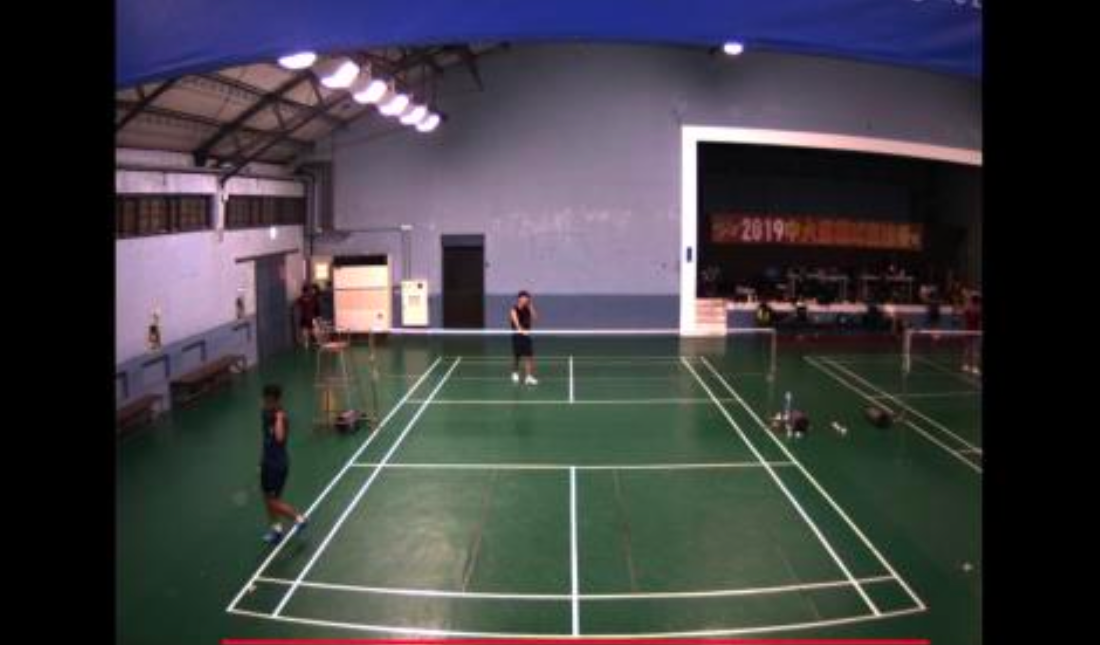}
        \caption{Pre distortion correction (Red line on the bottom does not align with the baseline).}
		\label{fig:pre_video_distortion_correction}
	\end{subfigure}
	\begin{subfigure}[t]{0.8\linewidth}
		\centering
		\includegraphics[width=1.0\linewidth]{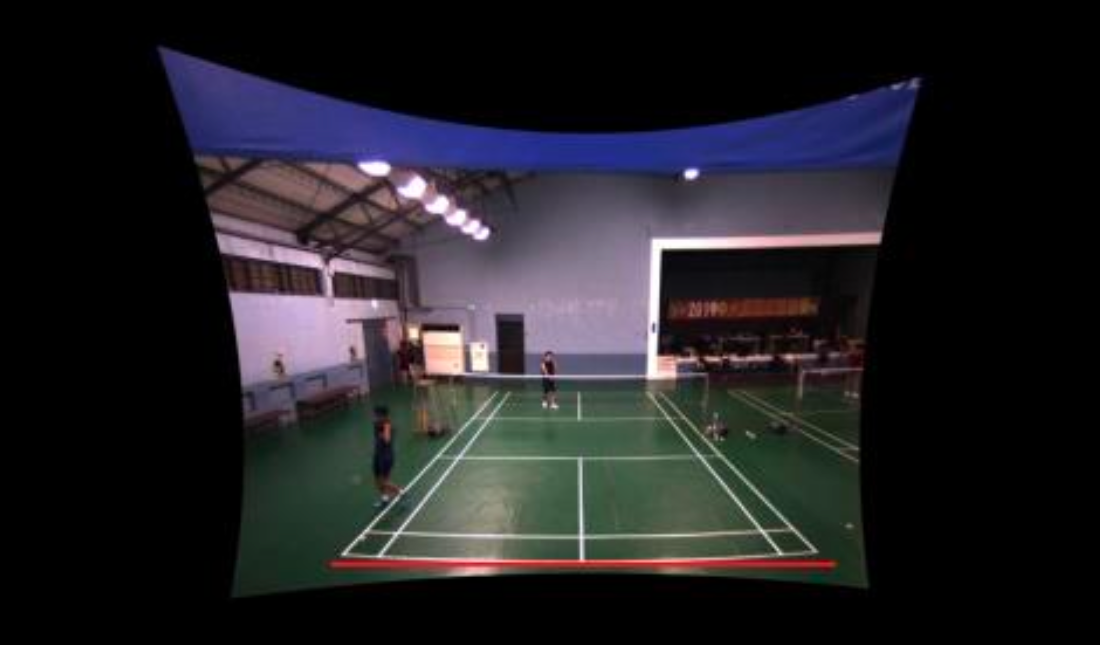}
        \caption{Post distortion correction (Red line on the bottom aligns with the baseline).}
		\label{fig:post_video_distortion_correction}
	\end{subfigure}
	\begin{subfigure}[t]{0.8\linewidth}
		\centering
		\includegraphics[width=1.0\linewidth]{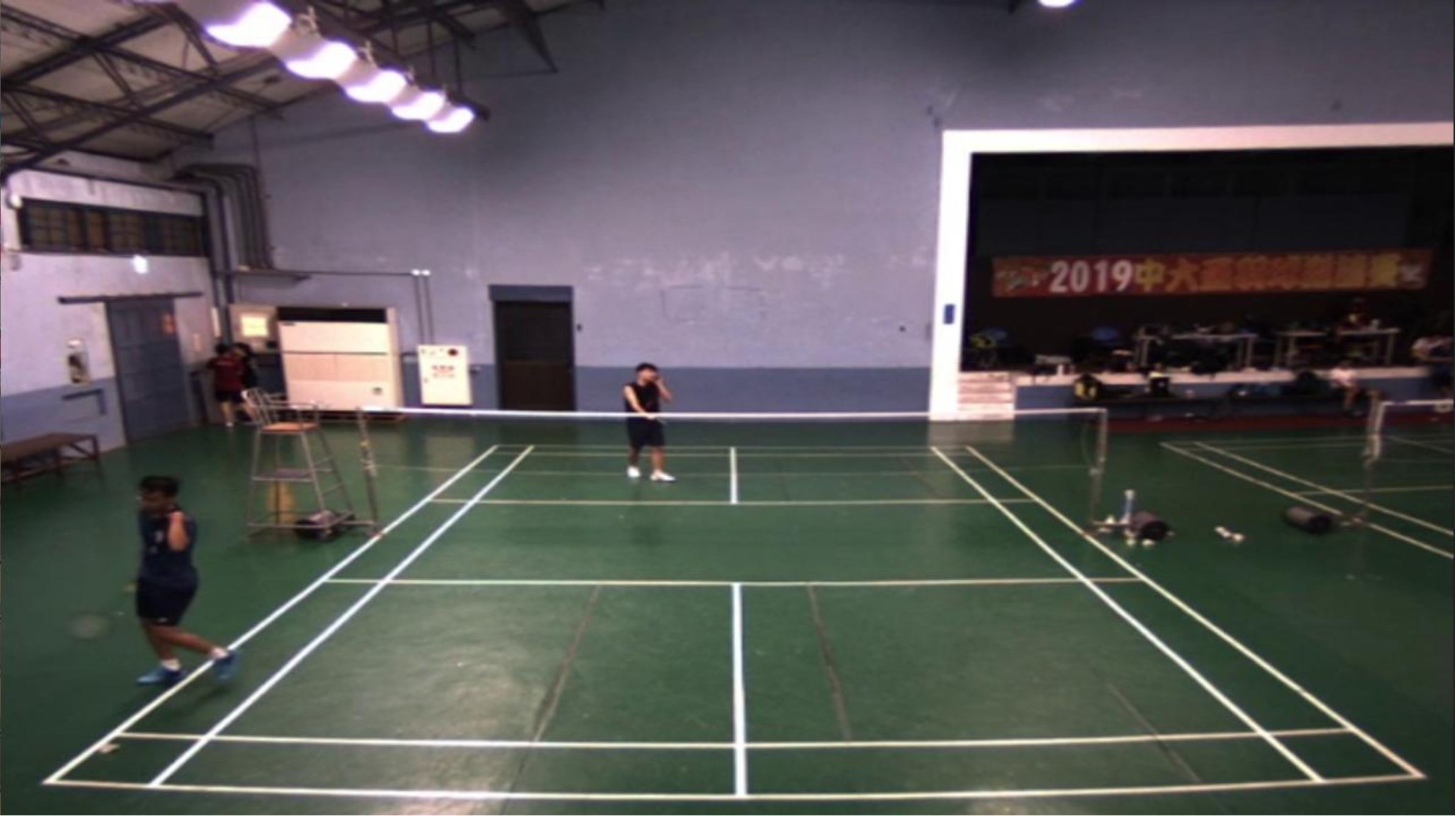}
        \caption{Crop and resize the post-distortion correction video.}
		\label{fig:crop_resize_video}
	\end{subfigure}
	\caption{Video distortion correction process.}
	\label{fig:video_preproces}
\end{figure}

Before the video correction, the originally straight boundary lines of the court appeared significantly curved under the wide-angle lens, which could cause errors in judging the court boundary lines and further lead to incorrect player positions and shuttlecock landing positions. After correction, the degree of curvature significantly improved compared to before correction. However, the screen was compressed due to the correction, so the necessary parts must be cut out and reprocessed into the required video size after correction. See the Figure \ref{fig:video_preproces}

\subsection{Human/Expert labeling}
Human labeling was a critical aspect of data processing. Five students participated in the labeling process, including a member of National Central University's badminton team and two from different badminton clubs. Their expertise and experience in the sport ensured accurate identification and labeling of the diverse badminton actions captured in the video data.

The labeled data underwent meticulous review by the head coach of National Central University's badminton team. The coach diligently examined the annotations, addressing any discrepancies or errors encountered during the labeling process. This rigorous review process ensured the reliability and validity of the labeled data, minimizing potential errors.

\subsection{Labeling process}
\begin{figure*}[ht]
\centering
\includegraphics[width=1.\textwidth]{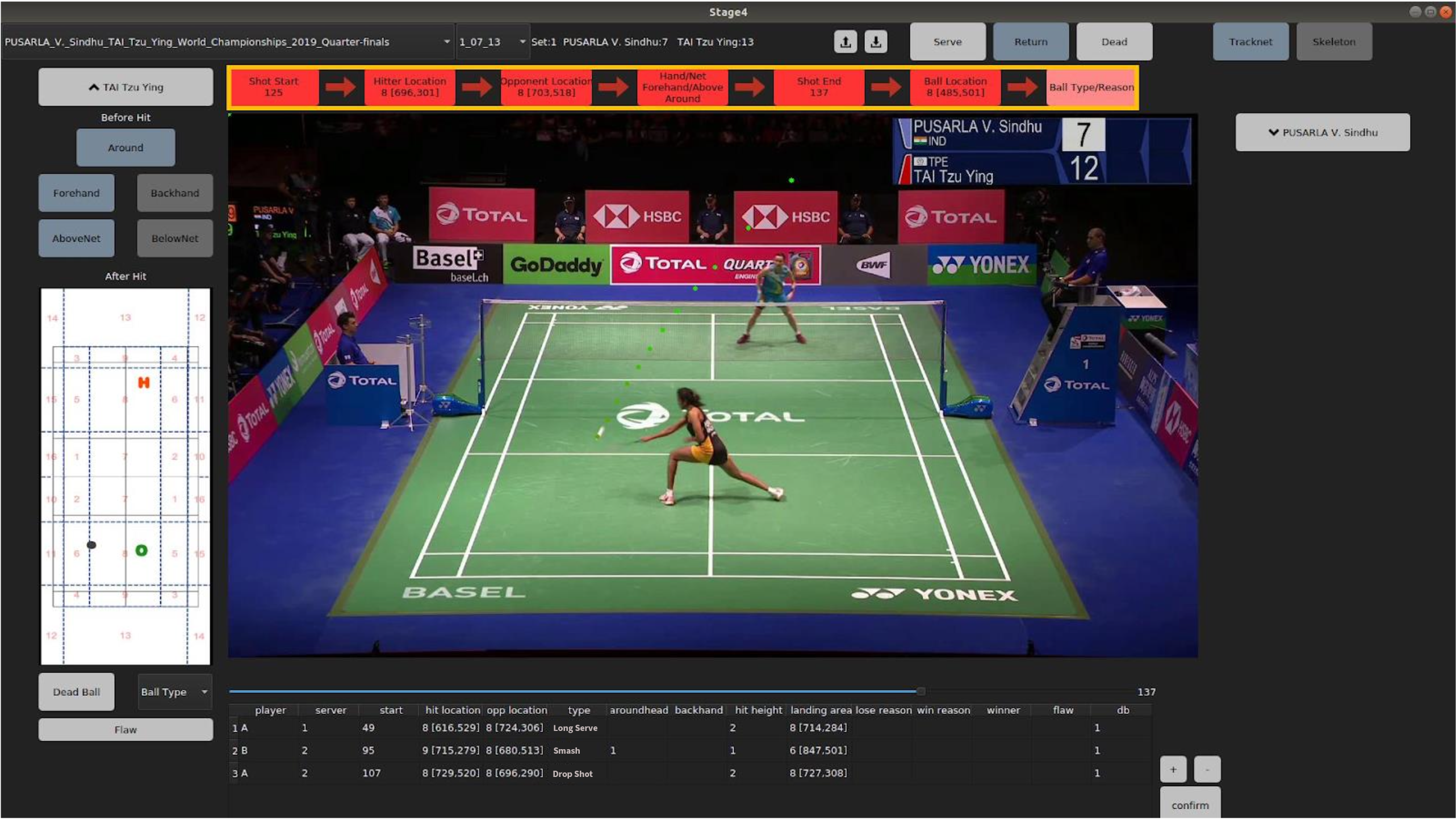}
\caption{The user interface of $S^2$ Labeling tool.}
\label{fig:S2_interface}
\end{figure*}

We use Shot-By-Shot ($S^2$) Labeling introduced in~\cite{huang2022s} as our main labeling tool. This tool consists of three parts: recording basic game data, cropping the video for each round, recording the score, and labeling micro-level game data. The interface of $S^2$ Labeling is shown in Figure \ref{fig:S2_interface}.
The process of labeling a badminton match video involves four stages. The first stage involves obtaining the match video and recording fundamental details about the competition. In the second stage, the beginning, end, and score of each rally are marked. Stage three focuses on pre-processing the video, which includes detecting the shuttlecock's trajectory, to aid in the microscopic labeling that follows in stage four, where each shot is analyzed in detail.

\subsection{Data post-processing}
Upon completing the data labeling, our research embarked on the task of segmenting the competition's content, which consisted of two primary areas. First, we addressed video segmenting: the full-length match videos were strategically divided into individual clips corresponding to specific ball types. This segmentation was executed post-data annotation, enhancing its applicability in video categorization or action recognition tasks. The slicing was performed based on FFmpeg, following the use of the Shot-By-Shot ($S^2$) Labeling tool, with segmentation guided by the labeled CSV file generated by this tool. File naming followed a convention encapsulating essential details like date, time, ID, start and end timestamps, halves of the game, and types of balls. Second, our data augmentation process filled gaps where some actions were less commonly found in actual gameplay. We enriched the dataset by employing controlled ball-feeding techniques and utilized the previously described camera setup to capture the diversity and complexity of gameplay actions. These combined efforts serve to create a robust and nuanced dataset, vital for our analytical and recognition tasks within the context of the sport.

\begin{figure}[ht]
\centering
\includegraphics[width=0.5\textwidth]{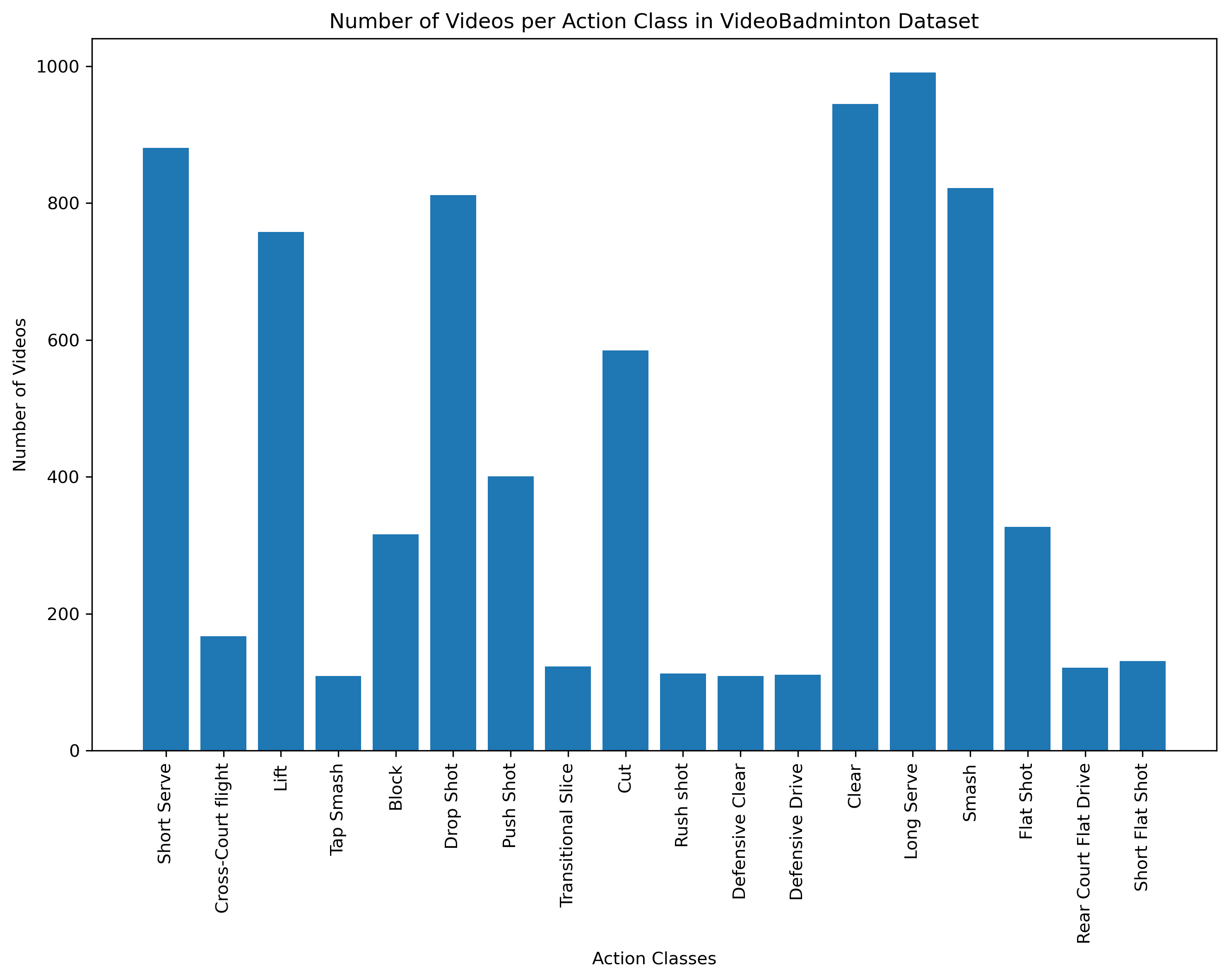}
\caption{Distribution of the number of action instances and videos per class.
}
\label{fig:sample_stats}
\end{figure}

\subsection{Dataset statistics and property}

To comprehensively analyze the VideoBadminton dataset, we employed both standard data analysis methods (see Figure \ref{fig:sample_stats}) and advanced techniques, including the calculation of frame entropy and the mean difference in frame-level features. This dual approach enhanced our understanding and characterization of the video content.

\subsubsection{Entropy of Video Frames}
The frame entropy measures the randomness or complexity within individual videos. Higher entropy suggests more complex and information-rich frames, whereas lower entropy indicates uniform and less detailed content. Computing the entropy of frames in a video provides insights into the variation and richness of visual information, which is essential in tasks like anomaly detection, content summarization, and activity recognition.

To compute the entropy of video frames, we employ the following steps:

\textbf{Conversion to Grayscale:} This step is implied in the formula by considering the frame as already being a grayscale image.
    
\textbf{Histogram Computation:} The histogram of pixel intensities is calculated in a grayscale image, where pixel intensities range from 0 to 255. Let $h(i)$ be the histogram count for pixel intensity $i$, where $i$ ranges from 0 to 255.
    
\textbf{Normalization of Histogram:} The histogram is normalized by dividing each $h(i)$ by the total number of pixels $N$. Let 
    \begin{equation}
        p(i) = \frac{h(i)}{N}
    \end{equation}
    be the normalized histogram.
    
\textbf{Entropy Computation:} The entropy of the histogram is then computed. Entropy, in information theory, is given by the formula 
    \begin{equation}
        -\sum_{i=0}^{255} p(i) \log_2 p(i)
    \end{equation}
    where the sum is over all possible pixel intensities. The corresponding LaTeX formula is:
    \begin{equation}
        \text{Entropy} = -\sum_{i=0}^{255} p(i) \log_2 p(i) \quad \text{where} \quad p(i) = \frac{h(i)}{N}
    \end{equation}

where $p(i)$ represents the probability of occurrence of pixel intensity $i$, $h(i)$ is the histogram count for pixel intensity $i$, and $N$ is the total number of pixels in the grayscale image. Note that in the case of a normalized histogram, $\sum_{i=0}^{255} p(i) = 1$.

\subsubsection{Mean Difference of Frame-level Features}
The mean difference of features between consecutive frames captures the temporal dynamics of a video. It measures how much the content changes from one frame to the next, which is crucial for recognizing actions and understanding motion patterns. By analyzing the average feature difference, we can gauge the speed and intensity of activities within the video, distinguishing between fast-paced, dynamic scenes and more static, slow-moving content.

Consider a pretrained ResNet-50 model, denoted as $F$. This model, when applied to an input image $I$, typically outputs a classification result. However, for the purpose of feature extraction, we modify $F$ to obtain an intermediate representation $\mathbf{f}$ from the layer before the last layer. This modified model, denoted as $F'$, is defined as follows:

\begin{equation}
    \mathbf{f} = F'(I)
\end{equation}

Here, $\mathbf{f}$ is the feature vector extracted from the input image $I$ using the modified ResNet-50 model $F'$, where $F'$ represents the ResNet-50 architecture excluding its final classification layer. This feature vector $\mathbf{f}$ is used to compute feature distances or for other subsequent processing steps.

The function \ref{eq:euclidean_dist} computes the Euclidean distance between consecutive feature vectors in a list of features. If $\mathbf{f}_i$ represents the $i$-th feature vector in 'features list', the Euclidean distance $d$ between the $i$-th and $(i-1)$-th feature vectors is calculated as:

\begin{equation}
    d(\mathbf{f}_i, \mathbf{f}_{i-1}) = \sqrt{\sum_{j=1}^{n} (f_{ij} - f_{(i-1)j})^2}
    \label{eq:euclidean_dist}
\end{equation}

Here, $f_{ij}$ is the $j$-th component of the $i$-th feature vector, and $n$ is the number of components in each feature vector. This calculation is repeated for all consecutive pairs of feature vectors in the list.



\begin{figure}[ht]
\centering
\includegraphics[width=0.45\textwidth]{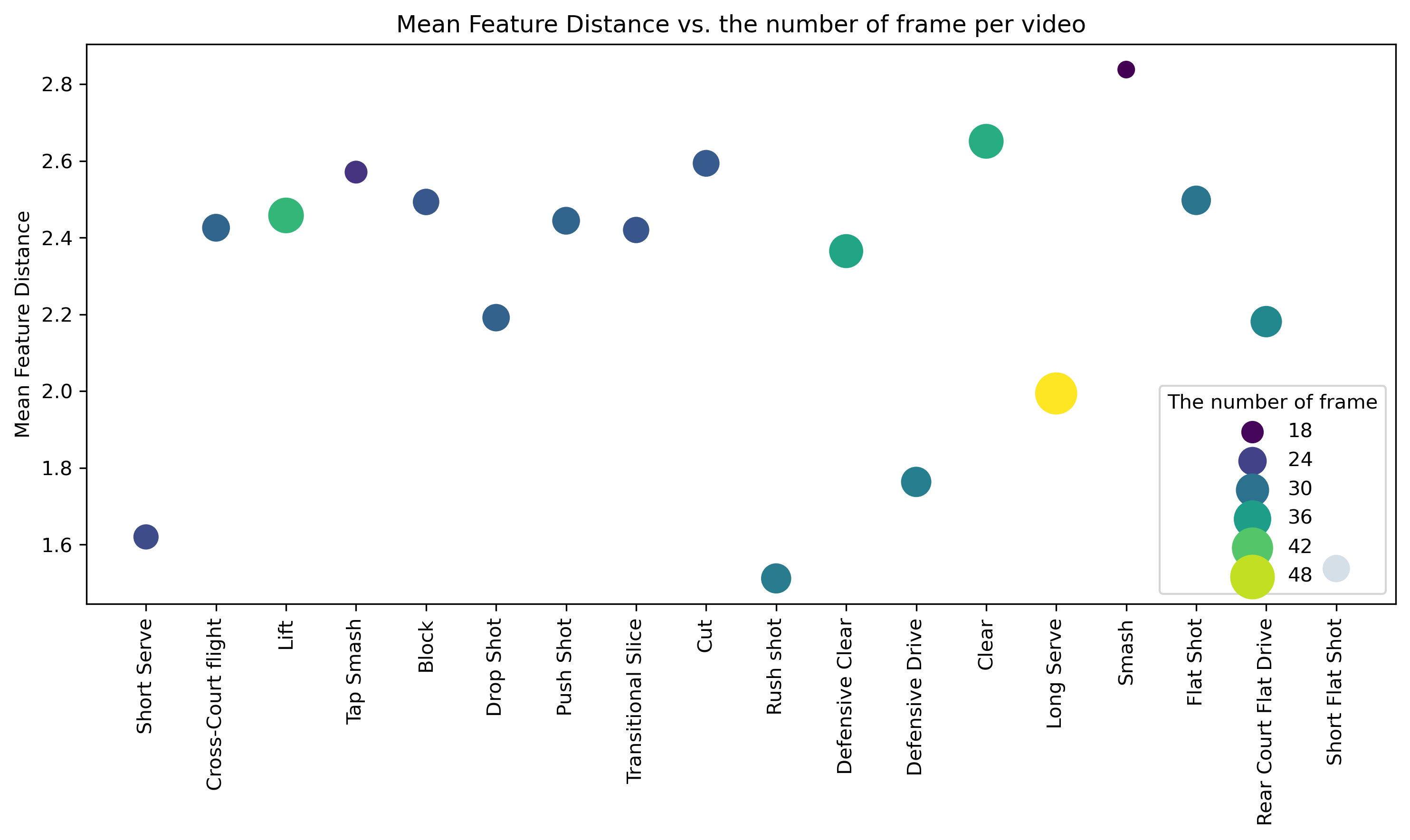}
\caption{Distribution of Mean Feature Distance by Class ID.}
\label{fig:feature_dist_vs_frame_num_fig}
\end{figure}
\begin{figure}[ht]
\centering
\includegraphics[width=0.45\textwidth]{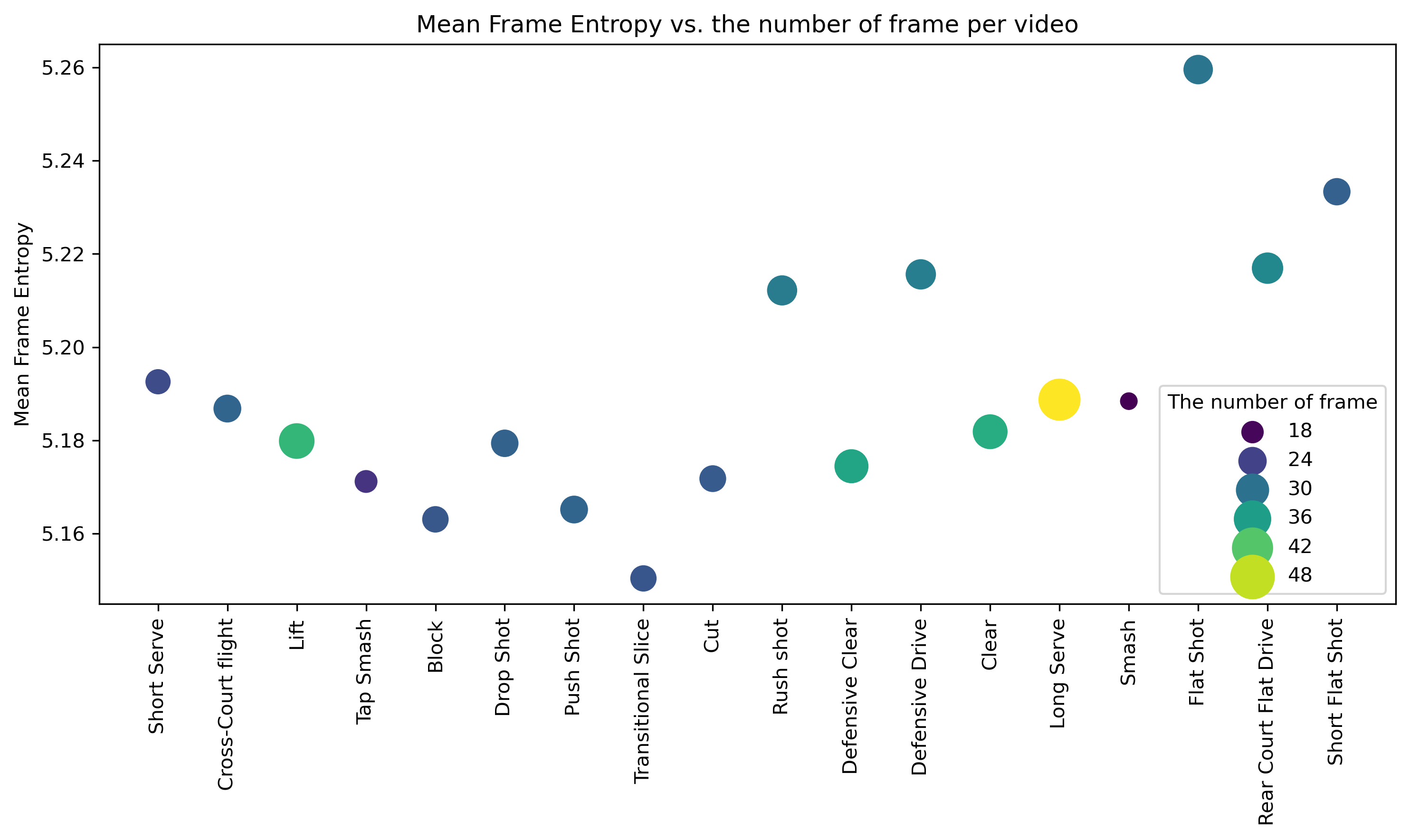}
\caption{Distribution of Mean Frame Entropy by Class ID.}
\label{fig:frame_entropy_vs_frame_num_fig}
\end{figure}
\begin{figure*}[ht]
\centering
\includegraphics[width=1.0\textwidth]{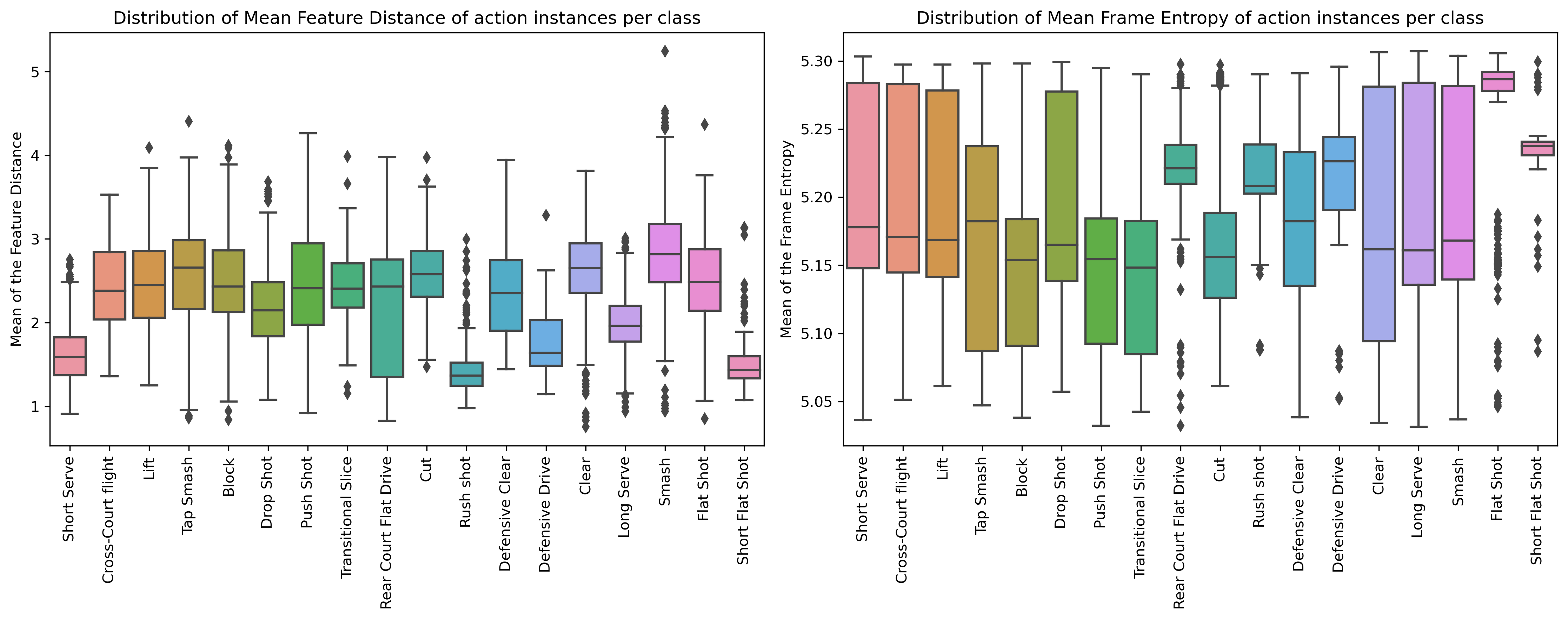}
\caption{Distribution of Mean Frame Entropy by Class ID.}
\label{fig:frame_entropy_and_feature_dist_fig}
\end{figure*}

The combination of frame entropy and mean feature difference (see Figures \ref{fig:feature_dist_vs_frame_num_fig}, \ref{fig:frame_entropy_vs_frame_num_fig}, \ref{fig:frame_entropy_and_feature_dist_fig}) offers a comprehensive toolset for analyzing video datasets. It enables us to understand both the spatial complexity and temporal progression of video content, thereby providing a robust foundation for advanced video processing and analysis techniques.

\section{Methods}

The selection of methods for our benchmark is comprehensive and representative of the most popular approaches in current action recognition research. By incorporating a diverse set of methodologies, ranging from video-based deep learning architectures to skeleton-based models, our study covers a wide spectrum of techniques prevalent in the field. Our results provide a holistic and detailed understanding of badminton action recognition.

The inclusion of various types of methods, such as  R(2+1)D\shortcite{tran2018closer}, SlowFast\shortcite{feichtenhofer2019slowfast}, TimeSformer\shortcite{bertasius2021space}, Swim\shortcite{liu2022video}, and MViT-V2\shortcite{li2022mvitv2} for their state-of-the-art performance in video-based action recognition, and two skeleton-based methods, ST-GCN\shortcite{yan2018spatial} and PoseC3D\shortcite{duan2022revisiting}, allows us to evaluate and compare different approaches to action recognition. This comparison is particularly valuable in understanding the strengths and weaknesses of each method when applied to the specific challenges presented by badminton action recognition.

\subsection{"R(2+1)D" spatiotemporal convolutional block}
The "R(2+1)D" spatiotemporal convolutional block is an approach that factorizes 3D convolutions into a combination of 2D spatial and 1D temporal convolutions. This method presents two primary advantages: increased model complexity without additional parameters and ease of optimization.
(2+1)D convolutions incorporate an extra nonlinear rectification between the spatial and temporal components. This effectively doubles the number of nonlinearities compared to traditional 3D convolutions, allowing the network to represent more complex functions without increasing the parameter count.
The decomposition of 3D convolutions into separate spatial and temporal operations results in lower training and testing losses. This indicates that the (2+1)D blocks are simpler to optimize than the intertwined spatiotemporal filters in full 3D convolutions.

\subsection{Spatial Temporal Graph Convolutional Network (ST-GCN)}
Spatial-Temporal Graph Convolutional Network (ST-GCN) is for skeleton-based action recognition. The ST-GCN model operates on a spatial-temporal graph of a skeleton sequence, where the body joints are represented as nodes and the natural connections between them are defined as edges. The model uses graph convolutional networks to learn the spatial patterns and temporal convolutional networks to learn the temporal patterns.

\subsubsection{Spatial Graph Convolutional Neural Network}
The convolution operation on a single frame is defined as:
   \begin{equation}
   f_{\text{out}}(x) = \sum_{h=1}^{K}\sum_{w=1}^{K} f_{\text{in}}(p(x, h, w)) \cdot w(h, w)
   \end{equation}
Here, \( f_{\text{out}} \) and \( f_{\text{in}} \) are the output and input feature maps, respectively. \( p \) is a sampling function mapping the spatial coordinates, and \( w \) is a weight function.

\subsubsection{Spatial Graph Convolution}

The graph convolution reformulated for the spatial domain is:
   \begin{equation}
   f_{\text{out}}(v_{ti}) = \sum_{v_{tj} \in B(v_{ti})} \frac{1}{Z_{ti}(v_{tj})} f_{\text{in}}(p(v_{ti}, v_{tj})) \cdot w(v_{ti}, v_{tj})
   \end{equation}
In this equation, \( B(v_{ti}) \) denotes the neighbor set of node \( v_{ti} \), \( Z_{ti}(v_{tj}) \) is a normalizing term, \( f_{\text{in}} \) and \( f_{\text{out}} \) are the input and output feature maps, and \( w \) represents the weight function.




\subsection{TimeSformer}
TimeSformer is a novel approach for video classification that leverages self-attention mechanisms, eliminating the need for convolutions. This method adapts the Transformer architecture to analyze videos by learning spatiotemporal features from sequences of frame-level patches. 
Given a set of input feature vectors $X = \{x_1, x_2, ..., x_n\}$, where $x_i \in \mathbb{R}^d$, the TimeSformer model computes a set of output vectors $Y = \{y_1, y_2, ..., y_n\}$, where $y_i \in \mathbb{R}^d$:
\begin{enumerate}
    \item Divide the input frames into non-overlapping segments of length $F$:
    \begin{equation}
        X = \{x_{1:F}, x_{F+1:2F}, ..., x_{n-F+1:n}\}.
    \end{equation}

    \item Apply self-attention within each segment and across segments:
    \begin{equation}
        \begin{aligned}
            Q &= XW_Q, K &= XW_K, V &= XW_V, \\
            \alpha_{i,j} &= \text{softmax}\left(\frac{Q_{i,j}K_{i,j}^T}{\sqrt{d}}\right), \\
            y_{i,j} &= \sum_{k=1}^n \alpha_{i,j,k}V_{j,k},
        \end{aligned}
    \end{equation}
    where $W_Q, W_K, W_V \in \mathbb{R}^{d \times d}$ are learnable weight matrices, $\alpha_{i,j}$ is a matrix of attention weights for the $i$-th query vector and the $j$-th key vector, and $\text{softmax}$ is the softmax function.

    \item Apply a learnable temporal position embedding to encode the order of frames within each segment:
    \begin{equation}
        X_{i,j} = X_{i,j} + P_j,
    \end{equation}
    where $P_j \in \mathbb{R}^d$ is a learnable position embedding for the $j$-th frame within each segment.

    \item Average the segment-level predictions and pass them through a fully connected layer to obtain the final classification:
    \begin{equation}
        \begin{aligned}
            z_i &= \frac{1}{F}\sum_{j=1}^F y_{i,j}, \\
            \hat{y}_i &= \text{softmax}(W_oz_i + b_o),
        \end{aligned}
    \end{equation}
    where $z_i \in \mathbb{R}^d$ is the segment-level prediction for the $i$-th segment, $W_o \in \mathbb{R}^{c \times d}$ and $b_o \in \mathbb{R}^c$ are learnable weight and bias parameters for the fully connected layer, and $\hat{y}_i \in \mathbb{R}^c$ is the predicted class probabilities for the $i$-th segment. The final prediction for the entire video is obtained by averaging the segment-level predictions:
    \begin{equation}
        \hat{y} = \frac{1}{n-F+1}\sum_{i=1}^{n-F+1} \hat{y}_i,
    \end{equation}
    where $\hat{y} \in \mathbb{R}^c$ is the predicted class probabilities for the entire video.
\end{enumerate}

The TimeSformer model achieves state-of-the-art results on several action recognition benchmarks, has low training and inference cost, and can be applied to longer video clips.

\subsection{PoseConv3D}
The PoseConv3D proposes a new approach for skeleton-based action recognition. The method involves creating 2D heatmaps for each joint and limb in each video frame, and then stacking these heatmaps along the temporal dimension to create a 3D heatmap volume. The authors then apply 3D convolutional neural networks to this volume to classify the action.

The formula for creating a joint heatmap $J$ is:
\begin{equation}
    J_{kij} = e^{-\frac{D((i,j),j_k)^2}{2\sigma_j^2}},
\end{equation}
where $j_k$ is the k-th joint, $D$ is the distance from the point $(i,j)$ to the joint $j_k$, and $\sigma_j$ is a parameter that controls the spread of the Gaussian.

The formula for creating a limb heatmap $L$ is:
\begin{equation}
    L_{kij} = e^{-\frac{D((i,j),seg[ak,bk])^2}{2\sigma_l^2 \cdot \min(ca_k,cb_k)}},
\end{equation}
where $ak$ and $bk$ are the two joints connected by the k-th limb, $seg[ak,bk]$ is the line segment connecting these joints, $D$ is the distance from the point $(i,j)$ to this segment, $ca_k$ and $cb_k$ are the confidence scores of the two joints, and $\sigma_l$ is a parameter that controls the spread of the Gaussian.

\subsection{SlowFast Network}

The SlowFast network architecture for video recognition captures both spatial semantics and motion at fine temporal resolution. The Slow pathway operates at a low frame rate and captures spatial semantics, while the Fast pathway operates at a high frame rate and captures motion. The two pathways are fused together to achieve state-of-the-art accuracy on major video recognition benchmarks.

Let $T$ be the number of frames in a video, $C$ be the number of input channels, and $S$ be the spatial resolution of the input frames. The SlowFast network takes as input a video tensor of size $T \times C \times S \times S$ and outputs a classification score for each action class. The Slow pathway operates on a downsampled version of the input tensor, obtained by applying a temporal stride of $\alpha$. The Fast pathway operates on the original input tensor, with a temporal stride of $\beta = \frac{\alpha}{k}$ where $k$ is an integer. The output of the two pathways is fused together using a lateral connection and a fusion operation, either summation or concatenation. The final output is obtained by applying a fully connected layer to the fused tensor.

\subsection{Multiscale Vision Transformers (MViTv2)}

The Multiscale Vision Transformers (MViTv2) are designed for visual recognition tasks including image classification, object detection, and video classification. The improvements include shift-invariant positional embeddings and a residual pooling connection to compensate for the effect of pooling strides in attention computation. These upgrades lead to significantly better results compared to prior work. The paper also introduces a generic MViT architecture for object detection and video recognition and provides five concrete instantiations of MViTv2 in increasing complexity.

Here is the attention mechanism used in MViTv2:
\begin{equation}
    \text{Attention}(Q,K,V) = \text{softmax}\left(\frac{QK^T}{\sqrt{d_k}}\right)V
\end{equation}
where $Q$, $K$, and $V$ are the query, key, and value tensors, respectively, and $d_k$ is the dimension of the key vectors.

The computation of the attention weights between the query tensor $Q$ and the key tensor $K$, which are then used to weight the values in $V$. The softmax function is applied to the dot product of $Q$ and $K^T$ divided by the square root of $d_k$. The resulting attention weights are then multiplied by $V$ to obtain the output of the attention mechanism.

In addition to the pooling attention mechanism, the paper also introduces the concept of decomposed relative positional embeddings, which are computed as follows:
\begin{equation}
    \text{PE}_{i,j} = \begin{cases} 
    \sin\left(\frac{i}{10000^{2j/d}}\right) & \text{if } j \text{ is even} \\
    \cos\left(\frac{i}{10000^{2(j-1)/d}}\right) & \text{if } j \text{ is odd} 
    \end{cases}
\end{equation}
where $\text{PE}_{i,j}$ is the $i$-th element of the $j$-th positional embedding, $d$ is the dimension of the embeddings, and $j$ ranges from 0 to $\lfloor\frac{d}{2}\rfloor-1$. The positional embeddings are added to the input embeddings of the Transformer blocks to provide information about the relative positions of the input tokens.

\section{Empirical Studies}

We conduct a comprehensive evaluation of various action recognition methods for badminton videos. We also include a practical study on the most typical action recognition method using MMAction2~\cite{2020mmaction2}. Unless specified otherwise, all training protocols adhere to the original papers. Our primary objective is to gain a deeper understanding of fine-grained actions. 

\subsection{Evaluation metrics}

We follow the standard practices~\cite{duan2022revisiting} in action recognition to evaluate the performance of models. Our evaluation framework primarily incorporates Top-1 Accuracy and Mean Class Accuracy. These metrics are widely recognized for their effectiveness in assessing model performance in action recognition tasks. Additionally, we include Top-5 Accuracy as an auxiliary metric to offer a more comprehensive comparison across models.

\textbf{Top-1 Accuracy}: This metric, essential in our evaluation, measures the proportion of instances where the model's highest-confidence prediction aligns with the true label. It directly indicates the model's precision in accurately identifying the most probable action.

\textbf{Top-5 Accuracy}: As an additional measure, Top-5 Accuracy evaluates whether the true label is among the top five predictions made by the model. This metric provides a broader perspective on the model's predictive capacity, accommodating a wider array of potential correct predictions.

\textbf{Mean Class Accuracy}: Reflecting the ethos of balanced and equitable model assessment, Mean Class Accuracy calculates the average accuracy across various action categories. This metric is particularly vital for our study as it addresses potential class imbalances and ensures a uniform evaluation standard across all classes, thereby reflecting the model's overall effectiveness in recognizing diverse badminton actions.

\subsection{The experimental setup} 
Following a structured experimental setup, our VideoBadminton dataset was partitioned into training, validation, and test subsets, following an 8:1:1 distribution ratio. The training subset was the foundation for model training, while the validation subset was utilized for hyperparameter optimization. Performance evaluation and results are reported based on the test subset. 

Furthermore, to enhance the generalizability of our models and mitigate the risk of overfitting, we incorporated a series of data augmentation techniques. These included random cropping, horizontal flipping, and color jittering - strategies that introduce a degree of variability and robustness into the training process.

\subsubsection{R(2+1)D} model is trained with 180 epochs and validated every 5 epochs. It employs Stochastic Gradient Descent (SGD) as its optimizer. The learning rate (lr) is set at 0.002, and momentum is at 0.9, and weight decay is set at 1e-4. Gradient clipping is employed with a maximum norm of 40 and L2 norm type to prevent exploding gradients. A CosineAnnealingLR scheduler is used for adjusting the learning rate.

\subsubsection{SlowFast} model is trained with 128 epochs and validated every 5 epochs. It employs SGD as its optimizer. The lr is set at 0.02, and momentum is at 0.9, and weight decay is set at 1e-4. Gradient clipping is employed with a maximum norm of 40 and L2 norm type to prevent exploding gradients. Both LinearLR and CosineAnnealingLR schedulers are used.

\subsubsection{TimeSformer} model is trained with 15 epochs and validated every epoch. It employs SGD with Nesterov momentum enabled as its optimizer. The lr is set at 0.0001, and momentum is at 0.9, and weight decay is set at 1e-4. The weight decay of the backbone's $cls_token$, $pos_embed$, and $time_embed$ are set to zero. Gradient clipping is employed with a maximum norm of 40 and L2 norm type to prevent exploding gradients. The MultiStepLR scheduler which adjusts the learning rate at specified milestones is used.
\subsubsection{Swim} model is trained with 30 epochs and validated every 3 epochs. It employs AmpOptim as its optimizer and the lr is set at 0.0002. A combination of LinearLR and CosineAnnealingLR schedulers is employed.
\subsubsection{Mvit-V2} model is trained with 100 epochs and validated every 5 epochs. The base lr is set to 4e-4 and the AdamW is used as optimizer. Gradient clipping is employed with a maximum norm of 1 and L2 norm type. A two-phase learning rate scheduling strategy is employed: LinearLR scheduler increases the learning rate linearly from a factor of 0.01 over the first 30 epochs.
CosineAnnealingLR takes over from the 30th epoch, adjusting the learning rate following a cosine curve, with a minimum rate of $ base_lr / 100$ by the end of the 100th epoch.
\subsubsection{ST-GCN} model is trained with 128 epochs and validated every epoch. It employs SGD with Nesterov momentum as its optimizer. The lr is set at 0.1, and momentum is at 0.9, and weight decay is set at 5e-4. A CosineAnnealingLR scheduler is employed for adjusting the learning rate.
\subsubsection{PoseC3D} model is trained with 12 epochs and validated every 5 epochs. It employs SGD as its optimizer. The lr is set at 0.04, and momentum is at 0.9, and weight decay is set at 3e-4. Gradient clipping is employed with a maximum norm of 40 and L2 norm type to prevent exploding gradients. A CosineAnnealingLR schedulers is used.

\subsection{Fine-tuning video action recognition models with the VideoBadminton dataset}

Fine-tuning existing video action recognition models like R(2+1)D, SlowFast, TimeSformer, Swim, MViT-V2, ST-GCN, and PoseC3D with the VideoBadminton dataset involves several critical steps. This process adapts pre-trained models to the specific characteristics of the VideoBadminton dataset, which may include unique actions or camera perspectives specific to badminton games.

\paragraph{Model Selection and Modification:}
Choose a suitable pre-trained model as the starting point. Depending on the complexity and size of the VideoBadminton dataset, a model with a suitable capacity should be selected to avoid overfitting or underfitting. The model might require modifications in its architecture, such as adjusting the number of output classes to match the number of action classes in the VideoBadminton dataset.
\paragraph{Transfer Learning and Fine-tuning:}
Transfer learning is applied using weights from a model pre-trained on a large and diverse dataset, such as Kinetics or ImageNet. Fine-tuning involves training the model on the VideoBadminton dataset while keeping the initial layers frozen and only training the latter layers, which allows the model to adapt to the specifics of badminton actions. 
\paragraph{Hyperparameter Tuning:}
Optimize hyperparameters such as learning rate, batch size, and number of epochs. A lower learning rate is generally preferred for fine-tuning to make small adjustments to the pre-trained weights. 
\paragraph{Regularization and Optimization:}
Apply regularization techniques like dropout and weight decay to prevent overfitting. Use appropriate optimization algorithms like Adam or SGD with momentum for efficient training.
\paragraph{Evaluation and Iteration:}
After fine-tuning, evaluate the model on a validation set and iteratively adjust the training process based on performance metrics like accuracy and loss. Pay close attention to the model's ability to generalize and its performance on each class to ensure balanced learning.

\begin{table*}
\centering
\caption{The performance of the models trained on the subset of the VideoBadminton.}
\begin{tabular}{c|c|c|c|c|c|c}
\toprule
& \multicolumn{3}{c|}{VideoBadminton-10 (10 samples/class)} & \multicolumn{3}{c}{VideoBadminton-50 (50 samples/class)} \\
\midrule
Methods & Top1 Acc & Top5 Acc & Mean Cls Acc & Top1 Acc & Top5 Acc & Mean Cls Acc  \\
\midrule

R(2+1)D\shortcite{tran2018closer}  &  13.10\% & 42.37\% & 14.50\% & 40.84\% & 77.18\% & 35.35\%  \\
SlowFast\shortcite{feichtenhofer2019slowfast} & 12.79\% & 38.08\% & 9.97\% & 12.28\% & 49.44\% & 21.16\%  \\
TimeSformer\shortcite{bertasius2021space}  & 19.45\% & 54.25\% & 24.25\% & 45.45\% & 85.16\% & 42.68\% \\
Swim\shortcite{liu2022video}  & 19.86\% & 56.91\% & 23.45\% & 53.53\% & 87.51\% & 48.92\%  \\
MViT-V2\shortcite{li2022mvitv2}  & 13.10\%    & 30.81\%  &   9.52\%     & 12.69\%   &  21.49\%  &    5.56\%     \\
\hline
ST-GCN\shortcite{yan2018spatial} &  28.05\% & 68.58\% & 23.59\% & 60.70\% & 89.25\% & 54.86\%  \\
PoseC3D\shortcite{duan2022revisiting} & 23.03\% & 59.77\% & 21.18\% & 59.98\% & 89.87\% & 50.21\% \\
\bottomrule
\end{tabular}

\label{tab:models_ablation}
\end{table*}

\subsection{Detailed analysis of action recognition methods}

This analysis presents a comparison of various methods used in video action recognition, evaluated based on three key performance metrics: Top-1 Accuracy (Top1 Acc), Top-5 Accuracy (Top5 Acc), and Mean Class Accuracy (Mean Cls Acc). The methods compared include R(2+1)D[45], SlowFast[11], TimeSformer[1], Swim[26], MViT-V2[24], ST-GCN[49], and PoseC3D[10].

SlowFast[11] emerges as the standout method, showing exceptional precision in identifying the primary action with an 82.80\% Top1 Acc and consistent performance across multiple guesses with a 97.54\% Top5 Acc. Its Mean Cls Acc of 73.80\% is the highest, indicating robustness and versatility in handling various action classes.

Conversely, MViT-V2[24] displays significant limitations with notably lower scores across all metrics, suggesting major challenges in action recognition. Other methods like R(2+1)D[45], Swim[26], ST-GCN[49], and PoseC3D[10] show a balance between Top-1 and Top-5 accuracies, each displaying strengths and weaknesses in different aspects of video action recognition.

TimeSformer[1], despite lower Top1 Acc and Mean Cls Acc, manages a reasonably high Top5 Acc, suggesting better performance with broader predictions. These varied performances highlight the ongoing challenges in developing video action recognition models that are both precise and versatile, capable of accurately identifying actions and performing consistently across diverse action categories.

\subsection{The studies of two balanced subsets of VideoBadminton}

Table \ref{tab:models_ablation} shows the performance metrics of various models trained on two balanced subsets of VideoBadminton, focusing on two specially curated subsets: VideoBadminton-10 and VideoBadminton-50. These subsets were methodically created by randomly selecting 10 and 50 clips from each class within the VideoBadminton dataset. This selection strategy ensures uniform sample distribution across all classes in both subsets, facilitating a comprehensive and balanced evaluation of diverse architectural models. The array of models assessed includes R(2+1)D, SlowFast, TimeSformer, Swim, MViT-V2, ST-GCN, and PoseC3D. Their performance is quantified using three key metrics: Top-1 Accuracy (Top1 Acc), Top-5 Accuracy (Top5 Acc), and Mean Class Accuracy (Mean Cls Acc).

\subsubsection{Performance on VideoBadminton-10:}
ST-GCN demonstrates the highest Top1 and Top5 Accuracies, suggesting its robustness in smaller datasets. TimeSformer and Swim also perform well, indicating their effectiveness in recognizing a diverse range of actions with limited data. Conversely, R(2+1)D, SlowFast, and MViT-V2 show lower accuracy, highlighting a potential need for larger datasets for optimal performance.

\subsubsection{Performance on VideoBadminton-50:}
In this larger dataset, ST-GCN leads in all metrics, showing its scalability. Swim and TimeSformer follow closely, especially in Top1 Acc and Top5 Acc, indicating strong action recognition capabilities. MViT-V2, however, exhibits significantly lower performance, suggesting limitations in scaling with increased data.

\subsubsection{Overall Observations:}
Models like ST-GCN and Swim show impressive adaptability and robustness across both subsets. The variance in performance across different models underscores the importance of model architecture in handling video action recognition tasks, particularly in relation to dataset size. These results underscore the importance of sample size in training effective action recognition models. While some models like ST-GCN and PoseC3D demonstrate reasonable performance with fewer samples, the overall trend suggests that increasing the number of samples per class leads to significantly better model performance. This highlights the necessity of a sufficiently large and diverse dataset for training more accurate and reliable action recognition models.

\begin{table}
\centering
\caption{Comparison of Action Recognition Methods for Badminton.}
\begin{tabular}{c|c|c|c}
\toprule
\textbf{Methods} & \textbf{Top1 Acc} & \textbf{Top5 Acc} & \textbf{Mean Cls Acc} \\ 
\midrule
R(2+1)D\shortcite{tran2018closer} & 79.53\% & 96.11\% & 66.97\% \\
SlowFast\shortcite{feichtenhofer2019slowfast} & 82.80\% & 97.54\% & 73.80\% \\
TimeSformer\shortcite{bertasius2021space} & 73.18\% & 94.78\% & 57.70\% \\
Swim\shortcite{liu2022video} & 81.99\% & 96.52\% & 69.93\% \\
MViT-V2\shortcite{li2022mvitv2} & 14.23\% & 62.23\% & 10.76\% \\
\hline
ST-GCN\shortcite{yan2018spatial} & 74.41\% & 93.76\% & 61.44\% \\
PoseC3D\shortcite{duan2022revisiting} & 80.76\% & 96.01\% & 67.18\% \\
\bottomrule
\end{tabular}

\label{tab:badminton_methods}
\end{table}

\section{Potential applications of the VideoBadminton dataset}

The VideoBadminton dataset, with its rich collection of badminton actions and movements, opens up some potential applications, both within and beyond the realm of sports science. These applications include: \textbf{Enhanced Athlete Training and Performance Analysis:} Coaches and athletes can utilize the dataset to analyze and improve player techniques and strategies. By comparing player actions with the diverse examples in the dataset, nuanced aspects of performance, such as stroke precision and footwork, can be evaluated and enhanced. \textbf{Automated Sports Broadcasting:} The dataset can be instrumental in developing algorithms for automated event detection and highlight generation in badminton matches. This would enable broadcasters to create more engaging and informative content for viewers. \textbf{Injury Prevention and Rehabilitation:} By analyzing the detailed movements captured in the dataset, researchers can identify patterns that may lead to injuries. This information can be used to develop training programs that focus on injury prevention and aid in the rehabilitation of athletes. \textbf{Academic Research in Human Kinetics:} The dataset is a valuable resource for researchers in human kinetics and biomechanics, enabling detailed studies on human movement, posture, and coordination in badminton.

\section{Conclusion}
In conclusion, we propose a new dataset called VideoBadminton, which is focused on badminton videos and differs from existing action recognition datasets in various aspects. The dataset provides high-quality, action-centric data with consistent annotations across multiple semantic and temporal granularities. Additionally, the dataset offers diverse and informative action instances, providing a valuable resource for action recognition research in the context of badminton. Moreover, we have conducted empirical investigations using the VideoBadminton dataset and have identified new challenges for future research. These findings are expected to facilitate new advances in the field of action understanding, not only in the context of badminton but also in other areas where action recognition is a crucial task. The availability of the VideoBadminton dataset is expected to promote more accurate and robust human action recognition methods, leading to significant progress in the field of sports analysis.

\bibliographystyle{ACM-Reference-Format}
\bibliography{reference}

\clearpage

\newpage
\appendix
\section{Badminton Action Classes for Video Recognition}

This section details the selection of action classes for our dataset, adhering to the rules and standards set by the Badminton World Federation. These action classes are essential for training video recognition models in badminton:

\begin{itemize}
    \item "Short Serve" - A serve that just clears the net and lands near the front of the service court.
    \item "Cross-Court flight" - A shot that travels diagonally from one corner of the court to the other.
    \item "Lift" - A defensive shot from the front or midcourt to the back of the opponents’ court.
    \item "Tap Smash" - A less powerful smash played with more wrist action.
    \item "Block" - A simple and soft shot just over the net, often in response to a smash.
    \item "Drop Shot" - A shot that just clears the net and falls rapidly.
    \item "Push Shot" - A controlled shot that pushes the shuttlecock to the back of the court.
    \item "Transitional Slice" - A deceptive stroke that changes the shuttlecock’s direction.
    \item "Cut" - A shot with an angled racket face to create a spinning motion.
    \item "Rush shot" - A fast and aggressive shot towards the opponent.
    \item "Defensive Clear" - A high and deep shot to the back of the opponent's court.
    \item "Defensive Drive" - A flat and fast defensive shot.
    \item "Clear" - A high shot that travels to the back of the opponent’s court.
    \item "Long Serve" - A serve that targets the back of the service court.
    \item "Smash" - A powerful downward shot.
    \item "Flat Shot" - A shot played flat over the net.
    \item "Rear Court Flat Drive" - A flat drive shot from the back of the court.
    \item "Short Flat Shot" - A quick and flat shot played from the front or midcourt.
\end{itemize}

The inclusion of these action classes ensures a comprehensive and effective dataset for training video action recognition models specific to badminton, capturing the full range of movements and strategies used in the sport.

\section{The configuration settings for the models trained. }

We provide a comprehensive overview of the training process, including a detailed key for configuration settings employed within the MMaction2 framework. This benchmark serves as the foundation for our experimental setup, ensuring replicability and rigor in our methodology. By disclosing these specifics, we aim to offer transparency and facilitate further research in the field, allowing others to build upon our work with a clear understanding of the procedural nuances and technical parameters that underpin our findings.

\subsubsection{R(2+1)D}
The details of training a R(2+1)D model on the full VideoBadminton for the action recognition are shown in Table \ref{tab:conf_r2plus1d}.
\begin{table}[ht]
\centering
\begin{tabularx}{0.45\textwidth}{|l|X|}
\hline
\textbf{Component} & \textbf{Details} \\ \hline
Model & Recognizer3D \\ \hline
\multicolumn{2}{|l|}{\textbf{Backbone}} \\ \hline
type & ResNet2Plus1d \\ \hline
depth & 34 \\ \hline
pretrained & None \\ \hline
pretrained2d & False \\ \hline
norm\_eval & False \\ \hline
conv\_cfg & \{ Conv2plus1d\} \\ \hline
norm\_cfg & \{ SyncBN, requires\_grad: True, eps: 0.001\} \\ \hline
conv1\_kernel & (3, 7, 7) \\ \hline
conv1\_stride\_t & 1 \\ \hline
pool1\_stride\_t & 1 \\ \hline
inflate & (1, 1, 1, 1) \\ \hline
spatial\_strides & (1, 2, 2, 2) \\ \hline
temporal\_strides & (1, 2, 2, 2) \\ \hline
zero\_init\_residual & False \\ \hline
\multicolumn{2}{|l|}{\textbf{Classification Head (cls\_head)}} \\ \hline
type & I3DHead \\ \hline
num\_classes & 18 \\ \hline
in\_channels & 512 \\ \hline
spatial\_type & avg \\ \hline
dropout\_ratio & 0.5 \\ \hline
init\_std & 0.01 \\ \hline
average\_clips & prob \\ \hline
\multicolumn{2}{|l|}{\textbf{Data Preprocessor}} \\ \hline
type & ActionDataPreprocessor \\ \hline
mean & [123.675, 116.28, 103.53] \\ \hline
std & [58.395, 57.12, 57.375] \\ \hline
format\_shape & NCTHW \\ \hline
\multicolumn{2}{|l|}{\textbf{Train Pipeline}} \\ \hline
\textbf{Step} & \textbf{Details} \\ \hline
1 &  DecordInit, io\_backend: disk \\ \hline
2 &  SampleFrames, clip\_len: 8, frame\_interval: 8, num\_clips: 1 \\ \hline
3 &  DecordDecode \\ \hline
4 &  Resize, scale: (-1, 256) \\ \hline
5 &  RandomResizedCrop \\ \hline
6 &  Resize, scale: (224, 224), keep\_ratio: False \\ \hline
7 &  Flip, flip\_ratio: 0.5 \\ \hline
8 &  FormatShape, input\_format: NCTHW \\ \hline
9 &  PackActionInputs \\ \hline
\end{tabularx}
\caption{Configuration Settings for Model R(2+1)D Training.}
\label{tab:conf_r2plus1d}
\end{table}

\clearpage
\subsubsection{SlowFast} 
The details of training a SlowFast model on the full VideoBadminton for the action recognition are shown in Tables \ref{tab:config_slowfast_a} and \ref{tab:config_slowfast_b}.
\begin{table}[ht]
\centering
\begin{tabularx}{0.5\textwidth}{|l|X|}
\hline
\textbf{Component} & \textbf{Details} \\ \hline
Model & Recognizer3D \\ \hline
\multicolumn{2}{|l|}{\textbf{Backbone}} \\ \hline
type & ResNet3dSlowFast \\ \hline
pretrained & None \\ \hline
resample\_rate & 8 \\ \hline
speed\_ratio & 8 \\ \hline
channel\_ratio & 8 \\ \hline
\multicolumn{2}{|l|}{\textbf{Slow Pathway}} \\ \hline
type & resnet3d \\ \hline
depth & 50 \\ \hline
pretrained & None \\ \hline
lateral & True \\ \hline
conv1\_kernel & (1, 7, 7) \\ \hline
dilations & (1, 1, 1, 1) \\ \hline
conv1\_stride\_t & 1 \\ \hline
pool1\_stride\_t & 1 \\ \hline
inflate & (0, 0, 1, 1) \\ \hline
norm\_eval & False \\ \hline
\multicolumn{2}{|l|}{\textbf{Fast Pathway}} \\ \hline
type & resnet3d \\ \hline
depth & 50 \\ \hline
pretrained & None \\ \hline
lateral & False \\ \hline
base\_channels & 8 \\ \hline
conv1\_kernel & (5, 7, 7) \\ \hline
conv1\_stride\_t & 1 \\ \hline
pool1\_stride\_t & 1 \\ \hline
norm\_eval & False \\ \hline
\multicolumn{2}{|l|}{\textbf{Classification Head (cls\_head)}} \\ \hline
type & SlowFastHead \\ \hline
in\_channels & 2304 \\ \hline
num\_classes & 18 \\ \hline
spatial\_type & avg \\ \hline
dropout\_ratio & 0.5 \\ \hline
average\_clips & prob \\ \hline
\multicolumn{2}{|l|}{\textbf{Data Preprocessor}} \\ \hline
type & ActionDataPreprocessor \\ \hline
mean & [123.675, 116.28, 103.53] \\ \hline
std & [58.395, 57.12, 57.375] \\ \hline
format\_shape & NCTHW \\ \hline
\end{tabularx}
\caption{Configuration Settings for Model SlowFast Training (a).}
\label{tab:config_slowfast_a}
\end{table}

\begin{table}[ht]
\centering
\begin{tabularx}{0.5\textwidth}{|l|X|}
\hline
\multicolumn{2}{|l|}{\textbf{Training Pipeline}} \\ \hline
\textbf{Step} & \textbf{Details} \\ \hline
1 &  DecordInit, io\_backend: disk \\ \hline
2 &  SampleFrames, clip\_len: 32, frame\_interval: 2, num\_clips: 1 \\ \hline
3 &  DecordDecode \\ \hline
4 &  Resize, scale: (-1, 256) \\ \hline
5 &  RandomResizedCrop \\ \hline
6 &  Resize, scale: (224, 224), keep\_ratio: False \\ \hline
7 &  Flip, flip\_ratio: 0.5 \\ \hline
8 &  FormatShape, input\_format: NCTHW \\ \hline
9 &  PackActionInputs \\ \hline
\end{tabularx}
\caption{Configuration Settings for Model SlowFast Training (b).}
\label{tab:config_slowfast_b}
\end{table}

\subsubsection{TimeSformer}
The details of training a TimeSformer model on the full VideoBadminton for the action recognition are shown in Table \ref{tab:config_timesformer}.
\begin{table}[ht]
\centering
\begin{tabularx}{0.5\textwidth}{|l|X|}
\hline
\textbf{Component} & \textbf{Details} \\ \hline
Model & Recognizer3D \\ \hline
\multicolumn{2}{|l|}{\textbf{Backbone}} \\ \hline
type & TimeSformer \\ \hline
num\_frames & 8 \\ \hline
img\_size & 224 \\ \hline
patch\_size & 16 \\ \hline
embed\_dims & 768 \\ \hline
in\_channels & 3 \\ \hline
dropout\_ratio & 0.0 \\ \hline
transformer\_layers & None \\ \hline
attention\_type & divided\_space\_time \\ \hline
norm\_cfg & \{ LN, eps: 1e-06\} \\ \hline
\multicolumn{2}{|l|}{\textbf{Classification Head (cls\_head)}} \\ \hline
type & TimeSformerHead \\ \hline
num\_classes & 18 \\ \hline
in\_channels & 768 \\ \hline
average\_clips & prob \\ \hline
\multicolumn{2}{|l|}{\textbf{Data Preprocessor}} \\ \hline
type & ActionDataPreprocessor \\ \hline
mean & [127.5, 127.5, 127.5] \\ \hline
std & [127.5, 127.5, 127.5] \\ \hline
format\_shape & NCTHW \\ \hline
\multicolumn{2}{|l|}{\textbf{Training Pipeline}} \\ \hline
\textbf{Step} & \textbf{Details} \\ \hline
1 &  DecordInit, io\_backend: disk \\ \hline
2 &  SampleFrames, clip\_len: 8, frame\_interval: 2, num\_clips: 1 \\ \hline
3 &  DecordDecode \\ \hline
4 &  RandomRescale, scale\_range: (256, 320) \\ \hline
5 &  RandomCrop, size: 224 \\ \hline
6 &  Flip, flip\_ratio: 0.5 \\ \hline
7 &  FormatShape, input\_format: NCTHW \\ \hline
8 &  PackActionInputs \\ \hline
\end{tabularx}
\caption{Configuration Settings for Model TimeSformer Training.}
\label{tab:config_timesformer}
\end{table}

\clearpage
\subsubsection{PoseC3D} 
The details of training a PoseC3D model on the full VideoBadminton for the action recognition are shown in Table \ref{tab:conf_posec3d}.
\begin{table}[ht]
\centering
\begin{tabularx}{0.5\textwidth}{|l|X|}
\hline
\textbf{Component} & \textbf{Details} \\ \hline
Model & Recognizer3D \\ \hline
\multicolumn{2}{|l|}{\textbf{Backbone}} \\ \hline
type & ResNet3dSlowOnly \\ \hline
depth & 50 \\ \hline
pretrained & None \\ \hline
in\_channels & 17 \\ \hline
base\_channels & 32 \\ \hline
num\_stages & 3 \\ \hline
out\_indices & (2, ) \\ \hline
stage\_blocks & (4, 6, 3) \\ \hline
conv1\_stride\_s & 1 \\ \hline
pool1\_stride\_s & 1 \\ \hline
inflate & (0, 1, 1) \\ \hline
spatial\_strides & (2, 2, 2) \\ \hline
temporal\_strides & (1, 1, 2) \\ \hline
dilations & (1, 1, 1) \\ \hline
\multicolumn{2}{|l|}{\textbf{Classification Head (cls\_head)}} \\ \hline
type & I3DHead \\ \hline
in\_channels & 512 \\ \hline
num\_classes & 18 \\ \hline
dropout\_ratio & 0.5 \\ \hline
average\_clips & prob \\ \hline
\multicolumn{2}{|l|}{\textbf{Training Pipeline}} \\ \hline
\textbf{Step} & \textbf{Details} \\ \hline
 1 &  UniformSampleFrames, clip\_len: 48 \\ \hline
 2 &  PoseDecode \\ \hline
 3 &  PoseCompact, hw\_ratio: 1.0, allow\_imgpad: True \\ \hline
 4 &  Resize, scale: (-1, 64) \\ \hline
 5 &  RandomResizedCrop, area\_range: (0.56, 1.0) \\ \hline
 6 & Resize, scale: (56, 56), keep\_ratio: False \\ \hline
 7 & \begin{tabular}[c]{@{}l@{}} Flip, flip\_ratio: 0.5, \\ left\_kp: [1, 3, 5, 7, 9, 11, 13, 15], \\ right\_kp: [2, 4, 6, 8, 10, 12, 14, 16]\end{tabular} \\ \hline
 8 & \begin{tabular}[c]{@{}l@{}} GeneratePoseTarget, sigma: 0.6, \\ use\_score: True, with\_kp: True, \\ with\_limb: False\end{tabular} \\ \hline
 9 & FormatShape, input\_format: NCTHW\_Heatmap \\ \hline
 10 &  PackActionInputs \\ \hline
\end{tabularx}
\caption{Configuration Settings for Model PoseC3D Training.}
\label{tab:conf_posec3d}
\end{table}

\subsubsection{Swim} 
The details of training a Swim model on the full VideoBadminton for the action recognition are shown in Table \ref{tab:conf_swim}.
\begin{table}[ht]
\centering
\begin{tabularx}{0.5\textwidth}{|l|X|}
\hline
\textbf{Component} & \textbf{Details} \\ \hline
Model & Recognizer3D \\ \hline
\multicolumn{2}{|l|}{\textbf{Backbone}} \\ \hline
type & SwinTransformer3D \\ \hline
arch & small \\ \hline
pretrained2d & True \\ \hline
patch\_size & (2, 4, 4) \\ \hline
window\_size & (8, 7, 7) \\ \hline
mlp\_ratio & 4.0 \\ \hline
qkv\_bias & True \\ \hline
qk\_scale & None \\ \hline
drop\_rate & 0.0 \\ \hline
attn\_drop\_rate & 0.0 \\ \hline
drop\_path\_rate & 0.2 \\ \hline
patch\_norm & True \\ \hline
\multicolumn{2}{|l|}{\textbf{Data Preprocessor}} \\ \hline
type & ActionDataPreprocessor \\ \hline
mean & [123.675, 116.28, 103.53] \\ \hline
std & [58.395, 57.12, 57.375] \\ \hline
format\_shape & NCTHW \\ \hline
\multicolumn{2}{|l|}{\textbf{Classification Head (cls\_head)}} \\ \hline
type & I3DHead \\ \hline
in\_channels & 768 \\ \hline
num\_classes & 18 \\ \hline
spatial\_type & avg \\ \hline
dropout\_ratio & 0.5 \\ \hline
average\_clips & prob \\ \hline
\multicolumn{2}{|l|}{\textbf{Training Pipeline}} \\ \hline
\textbf{Step} & \textbf{Details} \\ \hline
1 & type: DecordInit, io\_backend: disk \\ \hline
2 & type: SampleFrames, clip\_len: 32, frame\_interval: 2, num\_clips: 1 \\ \hline
3 & type: DecordDecode \\ \hline
4 & type: Resize, scale: (-1, 256) \\ \hline
5 & type: RandomResizedCrop \\ \hline
6 & type: Resize, scale: (224, 224), keep\_ratio: False \\ \hline
7 & type: Flip, flip\_ratio: 0.5 \\ \hline
8 & type: FormatShape, input\_format: NCTHW \\ \hline
9 & type: PackActionInputs \\ \hline
\end{tabularx}
\caption{Configuration Settings for Model Swim Training}
\label{tab:conf_swim}
\end{table}

\clearpage
\subsubsection{Mvit-V2}
The details of training a Mvit-V2 model on the full VideoBadminton for the action recognition are shown in Table \ref{tab:config_mvit-v2}.
\begin{table}[ht]
\centering
\begin{tabularx}{0.5\textwidth}{|l|X|}
\hline
\textbf{Component} & \textbf{Details} \\ \hline
Model & Recognizer3D \\ \hline
\multicolumn{2}{|l|}{\textbf{Backbone}} \\ \hline
type & MViT \\ \hline
arch & small \\ \hline
drop\_path\_rate & 0.2 \\ \hline
\multicolumn{2}{|l|}{\textbf{Data Preprocessor}} \\ \hline
type & ActionDataPreprocessor \\ \hline
mean & [114.75, 114.75, 114.75] \\ \hline
std & [57.375, 57.375, 57.375] \\ \hline
format\_shape & NCTHW \\ \hline
\multicolumn{2}{|l|}{Blending} \\ \hline
type & RandomBatchAugment \\ \hline
\multicolumn{2}{|l|}{Augments} \\ \hline
\ \ - type & MixupBlending, alpha: 0.8, num\_classes: 18 \\ \hline
\ \ - type & CutmixBlending, alpha: 1, num\_classes: 18 \\ \hline
\multicolumn{2}{|l|}{\textbf{Classification Head (cls\_head)}} \\ \hline
type & MViTHead \\ \hline
in\_channels & 768 \\ \hline
num\_classes & 18 \\ \hline
label\_smooth\_eps & 0.1 \\ \hline
average\_clips & prob \\ \hline
\multicolumn{2}{|l|}{\textbf{Training Pipeline}} \\ \hline
\textbf{Step} & \textbf{Details} \\ \hline
1 &  DecordInit, io\_backend: disk \\ \hline
2 &  SampleFrames, clip\_len: 16, frame\_interval: 4, num\_clips: 1 \\ \hline
3 &  DecordDecode \\ \hline
4 &  Resize, scale: (-1, 256) \\ \hline
5 &  PytorchVideoWrapper, op: RandAugment, magnitude: 7, num\_layers: 4 \\ \hline
6 &  RandomResizedCrop \\ \hline
7 &  Resize, scale: (224, 224), keep\_ratio: False \\ \hline
8 &  Flip, flip\_ratio: 0.5 \\ \hline
9 &  RandomErasing, erase\_prob: 0.25, mode: 'rand' \\ \hline
10 &  FormatShape, input\_format: NCTHW \\ \hline
11 &  PackActionInputs \\ \hline
\end{tabularx}
\caption{Configuration Settings for Model MViT-V2 Training.}
\label{tab:config_mvit-v2}
\end{table}

\subsubsection{ST-GCN}
The details of training an ST-GCN model on the full VideoBadminton for the action recognition are shown in Table \ref{tab:config_st-gcn}.
\begin{table}[ht]
\centering
\begin{tabularx}{0.5\textwidth}{|l|X|}
\hline
\textbf{Component} & \textbf{Details} \\ \hline
Model & RecognizerGCN \\ \hline
\multicolumn{2}{|l|}{\textbf{Backbone}} \\ \hline
type & STGCN \\ \hline
graph\_cfg & \{layout: 'coco', mode: 'stgcn\_spatial'\} \\ \hline
\multicolumn{2}{|l|}{\textbf{Classification Head (cls\_head)}} \\ \hline
type & GCNHead \\ \hline
num\_classes & 18 \\ \hline
in\_channels & 256 \\ \hline
\multicolumn{2}{|l|}{\textbf{Training Pipeline}} \\ \hline
\textbf{Step} & \textbf{Details} \\ \hline
1 & PreNormalize2D \\ \hline
2 & GenSkeFeat, dataset: 'coco', feats: ['b'] \\ \hline
3 & UniformSampleFrames, clip\_len: 100 \\ \hline
4 & PoseDecode \\ \hline
5 & FormatGCNInput, num\_person: 2 \\ \hline
6 & PackActionInputs \\ \hline
\end{tabularx}
\caption{Configuration Settings for Model ST-GCN Training.}
\label{tab:config_st-gcn}
\end{table}

\end{document}